\documentclass{IEEEoj}
\usepackage{cite}
\usepackage{amsmath,amssymb,amsfonts}
\usepackage{algorithmic}
\usepackage{graphicx,color}
\usepackage{textcomp}

\def\BibTeX{{\rm B\kern-.05em{\sc i\kern-.025em b}\kern-.08em
    T\kern-.1667em\lower.7ex\hbox{E}\kern-.125emX}}
\AtBeginDocument{\definecolor{ojcolor}{cmyk}{0.93,0.59,0.15,0.02}}
\def\OJlogo{\vspace{-4pt}\hskip-4pt\includegraphics[height=18pt]{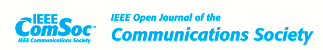}}

\usepackage{algorithm}
\usepackage{algorithmic}
\usepackage{comment}
\usepackage{multirow}
\usepackage{url}

\begin{document}
\receiveddate{XX Month, XXXX}
\reviseddate{XX Month, XXXX}
\accepteddate{XX Month, XXXX}
\publisheddate{XX Month, XXXX}
\currentdate{15 March, 2026}
\doiinfo{OJCOMS.2024.011100}

\title{Lightweight User-Personalization Method for Closed Split Computing}

\author{Yuya Okada\IEEEauthorrefmark{1}, Takyuki Nishio \IEEEauthorrefmark{1} \IEEEmembership{(Senior Member, IEEE)}}
\affil{School of Engineering, Institute of Science Tokyo, Tokyo 152-8550, Japan}
\corresp{CORRESPONDING AUTHOR: Takyuki Nishio (e-mail: nishio@ict.e.titech.ac.jp).}
\authornote{This work was supported in part by JST-ALCA-Next Japan Grant Number JPMJAN24F1 and JST PRESTO Grant Number JPMJPR2035.}
\markboth{Preparation of Papers for IEEE OPEN JOURNALS}{Author \textit{et al.}}

\begin{abstract}
\label{sec:abstract}
Split Computing enables collaborative inference between edge devices and the cloud by partitioning a deep neural network into an edge-side head and a server-side tail, reducing latency and limiting exposure of raw input data. However, in practical deployments, inference performance often degrades due to user-specific data distribution shifts, unreliable communication, and privacy-oriented perturbations. These challenges are further exacerbated in closed Split Computing environments where model architectures and parameters are inaccessible, making conventional fine-tuning infeasible.
To address this challenge, we propose SALT (Split-Adaptive Lightweight Tuning), a lightweight adaptation framework for closed Split Computing systems. SALT introduces a compact client-side adapter that refines intermediate representations produced by a frozen head network, enabling model adaptation without modifying the head or tail networks or increasing communication overhead. By modifying only the training conditions, SALT supports multiple adaptation objectives, including user personalization, communication robustness, and privacy-aware inference.
We evaluate SALT using ResNet-18 on CIFAR-10 and CIFAR-100 under three scenarios: user-specific data distributions, lossy communication environments, and noise-injected intermediate representations for privacy protection. Experimental results show that SALT achieves higher accuracy than conventional retraining and fine-tuning while significantly reducing training cost. On CIFAR-10, SALT improves personalized accuracy from $88.1\%$ to $93.8\%$ while reducing training latency by more than $60\%$. SALT also maintains over $90\%$ accuracy under $75\%$ packet loss and preserves high accuracy (approximately $88\%$ at $\sigma=1.0$) under noise injection.
These results demonstrate that SALT provides an efficient and practical adaptation framework for real-world Split Computing systems under closed-model constraints.
\end{abstract}


\maketitle

\section{INTRODUCTION}
% 推敲済み（20260122時点）
% ちょっと長いか...？

\label{sec:introduction}

Recent advances in deep neural networks have significantly improved the performance of cloud-based AI services.
However, executing large-scale models entirely on the cloud incurs high server-side computational costs, increased end-to-end latency, and potential privacy risks due to the transmission of raw input data.
To address these challenges, Split Computing has emerged as a promising paradigm,
in which a pre-trained neural network is partitioned into an edge device-side head network and a server-side tail network, enabling collaborative inference between edge devices and the cloud\cite{neurosurgeon, jointdnn}.
By processing part of the model on the edge device and transmitting only intermediate representations, Split Computing reduces latency while mitigating direct exposure of sensitive input data.
Comprehensive surveys further establish Split Computing as a fundamental architecture for efficient edge AI systems operating under resource and latency constraints\cite{sc_ee_survey}.

Beyond latency reduction, Split Computing has also been discussed as a building block for semantic communication and collaborative edge intelligence.
In this line of work, intermediate representations are regarded as task-relevant semantic features, and transmitting such features enables efficient inference under bandwidth-limited network conditions\cite{ib_task_oriented_comm, sec_semcom_survey}.

\begin{figure}[t]
    \centerline{\includegraphics[width=\linewidth]{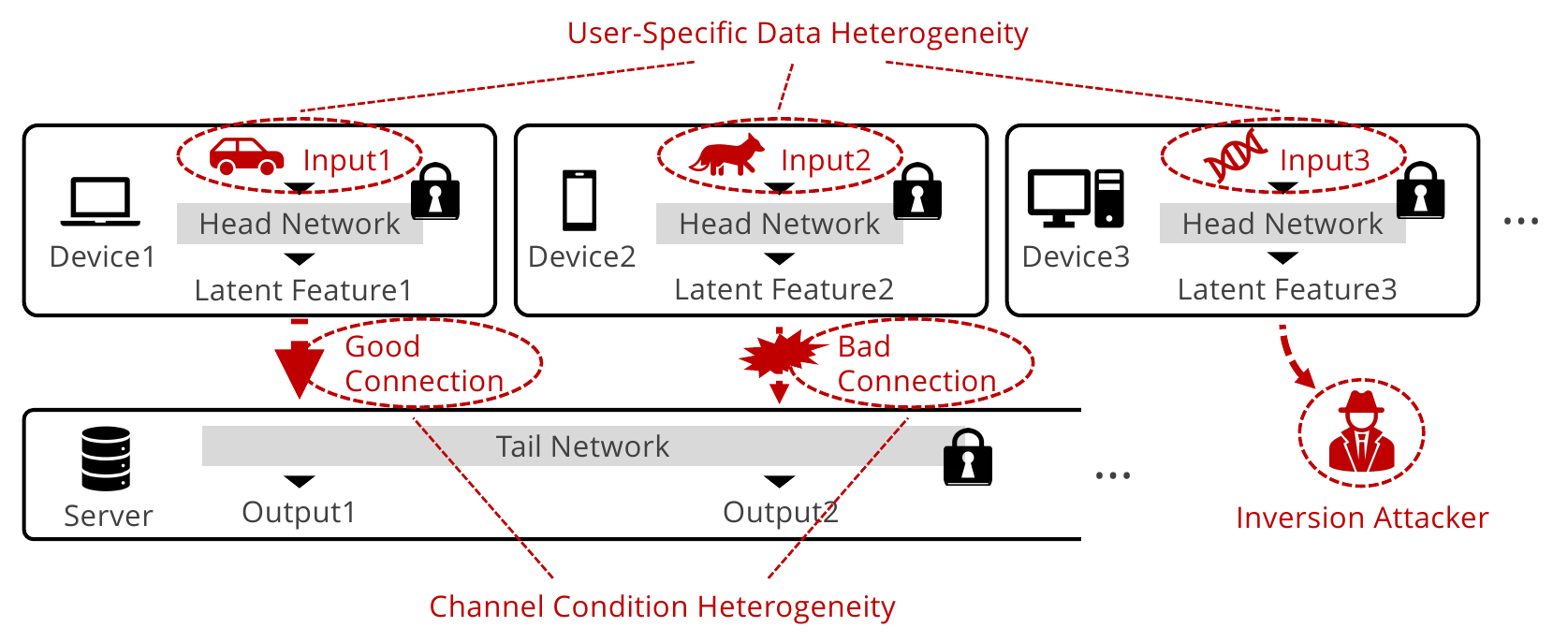}}
    \caption{Overview of practical challenges in Split Computing. Client devices execute a fixed head network and transmit intermediate representations to a server-side tail network under heterogeneous conditions. User-specific data distributions and varying communication quality can directly degrade inference performance, while transmitted representations may also be exposed to privacy threats such as inversion attacks in real-world edge--cloud systems.
}
    \label{fig:motivation}
\end{figure}

Despite these advantages, Split Computing is not achieved simply by partitioning a pre-trained model.
In practical deployments, inference performance often degrades significantly
unless the model is adapted to the target operating environment.
As illustrated in Fig.~\ref{fig:motivation}, real-world edge--cloud inference systems exhibit substantial heterogeneity.
Each client observes a user-specific data distribution, operates under distinct communication conditions, and transmits intermediate representations through potentially unreliable channels.
For example, users in urban environments may predominantly observe vehicles,
whereas users in rural environments may encounter animals more frequently.
In addition, packet loss and unstable wireless channel can severely distort transmitted latent features, leading to significant accuracy degradation if the model is not robust to communication impairments\cite{comtune}.

Furthermore, although Split Computing avoids direct transmission of raw input data, intermediate representations may still leak sensitive information.
Prior work on model inversion attacks demonstrates that inputs can be reconstructed from model outputs or latent representations under realistic assumptions\cite{fredrikson_mi_confidence}.
Recent studies further show that intermediate representations at the split layer can constitute a significant source of privacy leakage, and that such leakage can be quantitatively characterized\cite{refil_fisher_split_privacy}.
These results indicate that Split Computing is not inherently privacy-safe, and that robustness to intentional perturbations for privacy protection is an important practical requirement, for which noise injection into intermediate representations is often considered as a simple and practical measure.

Taken together, these observations suggest that model adaptation is almost indispensable in practical Split Computing systems.
In this study, we focus on three representative adaptation objectives:
(i) personalization to user-specific data distributions,
(ii) robustness against communication degradation such as packet loss, and
(iii) compensation for accuracy degradation caused by intentional noise injection
for privacy protection.

A common approach to such adaptation is conventional fine-tuning, where part of the model---typically the head network---is retrained or adjusted using task- or user-specific data.
Numerous studies have shown the effectiveness of fine-tuning for personalization and robustness to degraded communication environments\cite{tinytl, pockengine}.
However, these approaches implicitly assume that the internal architectures and parameters of the deployed models are accessible and modifiable.
In practice, this assumption often does not hold.
Many AI models are deployed as proprietary binaries, accessed through restricted APIs, or embedded in specialized hardware accelerators to protect intellectual property and ensure low-latency inference.
Recent studies on black-box and API-based adaptation further confirm that closed-model constraints are realistic rather than exceptional\cite{label_privacy_peft_split, aem_blackbox_domain_adapt}.
Under such conditions, conventional fine-tuning becomes difficult or impossible, creating a gap between existing adaptation methods and real-world Split Computing deployments.

To address this challenge, we propose SALT (Split-Adaptive Lightweight Tuning),
a lightweight model adaptation framework designed for closed Split Computing environments.
Instead of modifying the head or tail networks, SALT introduces a compact, trainable adapter on the client side that refines intermediate latent features produced by the frozen head network.
By estimating and adding a correction vector to the latent representation, SALT enables model adaptation without accessing or modifying the original models and without increasing communication overhead.
This design allows SALT to remain compatible with existing Split Computing architectures while operating under strict closed-model constraints.
Importantly, SALT provides a unified adaptation framework.
By changing only the training conditions while keeping the architecture unchanged, SALT can be applied to user-specific personalization, robust inference under communication degradation, and robustness to noise perturbations for privacy protection.

A preliminary version of SALT was presented at the \textit{IEEE GLOBECOM Workshop 2025}, where its effectiveness for personalization and packet-loss robustness
under closed-model constraints was demonstrated \cite{salt_arxiv}.
Building on this foundation, the present work discusses a unified adaptation perspective, showing that a single lightweight framework can adapt to diverse user conditions, and further extends the analysis by providing qualitative and visual insights into SALT’s robustness under degraded communication environments as well as privacy-aware inference evaluations under noise injection.

In this work, we evaluate SALT under three representative adaptation objectives:
adaptation to user-specific data distributions, adaptation to lossy communication environments, and adaptation to noise-injected intermediate representations for privacy-aware inference.
Experiments are conducted using ResNet-18 with the CIFAR-10 and CIFAR-100 datasets.
The results confirm the effectiveness of SALT for user-specific adaptation,
and its robustness under lossy communication and noise-injected conditions
in closed-model settings.

\vspace{5mm}
\noindent\textbf{Contributions:}
The main contributions of this work are summarized as follows:
\begin{itemize}
    %\item We formulate the problem of model adaptation in Split Computing under realistic closed-model constraints, considering data heterogeneity, communication degradation, and privacy-oriented perturbations. 
    \item We propose SALT, a lightweight and model-agnostic adaptation framework
    that enables effective adaptation without access to
    the parameters or internal structures of the head and tail networks.
    \item Through extensive experiments,
    we demonstrate that SALT achieves high inference accuracy
    with low training cost,
    maintains robustness under severe packet loss,
    and effectively compensates for accuracy degradation caused by noise injection
    while preserving resistance to inversion attacks.
\end{itemize}

The remainder of this work is organized as follows.
Section~\ref{sec:relatedwork} reviews related work.
Section~\ref{sec:method} presents the SALT framework.
Section~\ref{sec:evaluation} reports experimental results,
and Section~\ref{sec:conclusion} concludes the work.

\section{RELATED WORK}

\label{sec:relatedwork}

This section reviews prior studies on Split Computing from four perspectives:
system-level design, lightweight adaptation, communication robustness, and privacy protection.
While existing works largely assume access to model internals, this work focuses on closed split computing environments.
The following subsections position SALT relative to these prior approaches.

\subsection{Foundations of Split Computing}
\label{subsec:sc_foundation}

\subsubsection{System-Level Split Computing Frameworks}
The core concept of Split Computing was first formalized by Kang \textit{et al.} through Neurosurgeon, which demonstrated that dynamically partitioning a deep neural network at the layer level can significantly reduce inference latency and energy consumption under heterogeneous device and network conditions~\cite{neurosurgeon}.
This work established a key design principle of Split Computing: inference performance should be optimized by jointly considering computation placement and communication overhead.

Building on this principle, JointDNN extended Split Computing beyond inference to support both training and inference in a unified framework~\cite{jointdnn}.
By formulating layer-wise optimization problems that incorporate device-side constraints such as battery capacity and cloud-side workload, JointDNN highlighted that Split Computing is not merely a static partitioning strategy, but a comprehensive system-level framework that integrates computation, communication, and learning.

\subsubsection{Architecture Optimization and Automated Split Design}

Following early system-level studies, several works have explored how to optimize network architectures and split points to improve the latency--accuracy trade-off in Split Computing.
Neural architecture search (NAS) techniques have been introduced to automatically determine optimal split locations and network configurations under given hardware and network constraints~\cite{nas_for_split_computing,supernet_ft_nas_split}.
These studies demonstrate that treating split design as a parameterized optimization problem enables systematic exploration of performance trade-offs beyond manually selected architectures.

In contrast, practical Split Computing systems often operate under closed-model constraints, where both the head and tail networks are already deployed as fixed components and cannot be modified.
Under such conditions, architectural re-optimization or split redesign is infeasible, even when inference performance degrades due to changing operating environments.
This limitation highlights the need for post-deployment adaptation mechanisms that can improve inference performance without altering the original model architecture.

% Although these approaches focus on optimizing the structure of the main network, their underlying perspective---that split inference performance can be effectively controlled through parameterized architectural choices---is relevant to the design of lightweight adaptation modules.
% In particular, this line of work provides conceptual motivation for determining the structure and capacity of auxiliary components, such as the SALT adapter, in a principled and parameter-driven manner.

% ↓ 直接的な関連性がないため消した方が良い？
% \subsection{Communication-Efficient Intermediate Representation Transmission}
% Another important research direction in Split Computing focuses on reducing communication overhead by compressing intermediate representations.
% BottleNet++ introduced an encoder--decoder bottleneck architecture that explicitly compresses transmitted features while preserving end-to-end trainability~\cite{bottlenetpp}.
% Similarly, vector-quantized bottleneck approaches dynamically adjust the representation size to balance accuracy and latency under varying bandwidth conditions~\cite{vq_bottleneck_dynamic_split}.

% While these methods are effective in improving communication efficiency, they typically require architectural modifications to both the head and tail networks and joint retraining of the entire model.
% As a result, they implicitly assume that model structures and parameters are accessible and modifiable, which may not hold in many real-world Split Computing deployments.

\subsection{Lightweight Model Adaptation and Personalization}

A complementary research direction focuses on lightweight model adaptation and personalization, aiming to adapt pre-trained models to new tasks or domains with limited computational and memory overhead. 
TinyTL demonstrated that freezing backbone weights and updating only bias-related parameters can significantly reduce memory consumption for on-device learning \cite{tinytl}. 

In natural language processing, adapter-based methods insert small trainable modules into pre-trained models, enabling task-specific adaptation with a minimal number of parameters \cite{houlsby_adapters}. 
Similar ideas have been explored in computer vision, where residual adapters allow a single backbone to support multiple visual domains with high parameter sharing \cite{residual_adapters_visual_domains}. 
LoRA further reduced adaptation cost by injecting low-rank trainable matrices into frozen models, achieving performance comparable to full fine-tuning for large language models \cite{lora}. 

Beyond algorithmic design, PockEngine proposed a system-level approach for efficient fine-tuning on edge devices through sparse backpropagation and compile-time graph optimization \cite{pockengine}. 
Although these methods are effective under resource constraints, they inherently assume access to internal model parameters and structures.
As a result, they cannot be directly applied to Split Computing systems where the head and tail networks are deployed as closed or proprietary components.

\subsection{Communication Robustness in Distributed Inference}

In practical Split Computing systems, intermediate representations are transmitted over unreliable networks, making inference performance sensitive to packet loss and communication degradation. 
Several studies have addressed this issue by incorporating communication impairments into the training process.
Communication-oriented model fine-tuning simulates packet loss using dropout-like mechanisms, enabling models to maintain accuracy under highly lossy IoT networks \cite{comtune}. 

Similarly, packet-loss-tolerant split inference methods have been proposed to eliminate retransmissions while preserving inference accuracy, even under severe packet loss rates \cite{si_nr_packet_loss_tolerant_split}. 
Loss-Adapter further introduced a plug-and-play module that improves robustness by explicitly modeling packet loss distributions during training \cite{loss_adapter}. 

These approaches demonstrate that robustness to communication degradation can be learned; however, they typically rely on retraining or fine-tuning internal model parameters. 
This assumption limits their applicability in closed Split Computing environments, where head and tail networks cannot be modified after deployment.

\subsection{Privacy Risks and Protection in Split Computing}

Split Computing is generally considered beneficial for privacy, as it avoids direct transmission of raw input data; however, intermediate representations exchanged between devices and servers may still leak sensitive information about the input data.

Fredrikson \textit{et al.} demonstrated that model inversion attacks can reconstruct recognizable input data from model outputs or intermediate representations, revealing privacy vulnerabilities in neural networks \cite{fredrikson_mi_confidence}.
In the context of distributed and federated learning, it has also been shown that input data can be reconstructed from shared gradients \cite{inverting_gradients}, indicating that intermediate information exchanged during collaborative inference can serve as an attack surface.

To mitigate these risks, cryptography-based approaches have been proposed. 
For example, Delphi enables secure neural network inference by jointly designing cryptographic protocols and model architectures, allowing inference without revealing either user inputs or model parameters \cite{delphi_crypto_inference}.
While such approaches provide strong theoretical privacy guarantees, they incur substantial computational and communication overhead, which limits their practicality in real-time and resource-constrained edge--cloud scenarios.

As a lightweight and practical alternative, noise injection into intermediate representations has been widely adopted to reduce the reconstructability of sensitive information. By perturbing transmitted features, noise injection improves privacy while remaining compatible with existing Split Computing pipelines.
However, this approach inevitably degrades inference accuracy, resulting in a trade-off between privacy and utility.

Several studies have attempted to quantify and control this trade-off. 
NoPeek reduces information leakage by minimizing the correlation between input data and intermediate representations \cite{nopeek}. Other works proposed Fisher-information-based metrics to measure and bound privacy leakage in split-layer representations \cite{refil_fisher_split_privacy,dfil_bound_invertibility}.
Despite their effectiveness, many of these methods assume access to internal model parameters or require retraining of the head network, which is infeasible in closed Split Computing environments.

\subsection{Position of Our Work}
\label{subsec:position}

\begin{table*}[t]
\centering
\caption{Summary of related work discussed in this work and their assumptions in split computing.}
\label{tab:rw_compact}
\small
\setlength{\tabcolsep}{6pt}
\renewcommand{\arraystretch}{1.15}
\resizebox{\linewidth}{!}{%
\begin{tabular}{p{0.38\linewidth} p{0.22\linewidth} p{0.24\linewidth} p{0.16\linewidth}}
\hline
\textbf{Work} & \textbf{Primary Focus} & \textbf{Adaptation / Targeted Issue} & \textbf{Model Access Required} \\
\hline
Neurosurgeon \cite{neurosurgeon} 
& Split point scheduling 
& Latency / energy efficiency 
& Full \\
\hline
JointDNN \cite{jointdnn} 
& Split training/inference framework 
& Latency / energy efficiency 
& Full \\
\hline
Split Computing \& Early Exiting Survey \cite{sc_ee_survey} 
& Survey / challenges 
& Research landscape 
& -- \\
\hline
TinyTL \cite{tinytl} 
& Memory-efficient fine-tuning 
& Task / domain shift 
& Full \\
\hline
Adapters (NLP) \cite{houlsby_adapters} 
& Parameter-efficient transfer 
& Task adaptation 
& Full \\
\hline
Residual Adapters \cite{residual_adapters_visual_domains} 
& Multi-domain adaptation 
& Domain shift 
& Full \\
\hline
LoRA \cite{lora} 
& Low-rank adaptation 
& Task / domain shift 
& Full \\
\hline
PockEngine \cite{pockengine} 
& On-device tuning system 
& Training efficiency (edge) 
& Full \\
\hline
COMtune \cite{comtune} 
& Communication-aware tuning 
& Packet loss robustness 
& Full \\
\hline
SI-NR \cite{si_nr_packet_loss_tolerant_split} 
& Packet-loss-tolerant split inference 
& Packet loss robustness 
& Full \\
\hline
Loss-Adapter \cite{loss_adapter} 
& Plug-in robustness module 
& Packet loss robustness 
& Partial \\
\hline
Model inversion \cite{fredrikson_mi_confidence} 
& Privacy attack 
& Input reconstruction risk 
& None \\
\hline
Inverting gradients \cite{inverting_gradients} 
& Privacy attack 
& Gradient leakage risk 
& Full \\
\hline
NoPeek \cite{nopeek} 
& Privacy-aware training 
& Leakage reduction (training-time) 
& Full \\
\hline
ReFIL \cite{refil_fisher_split_privacy} 
& Privacy metric/control 
& Split-layer leakage control 
& Full \\
\hline
Delphi \cite{delphi_crypto_inference} 
& Cryptographic inference 
& Strong privacy guarantees 
& Protocol-level \\
\hline
\textbf{SALT (ours)}  
& Closed-SC adaptation 
& \textbf{Data shift / Communication degradation / Privacy-induced distortion} 
& \textbf{None} \\
\hline
\end{tabular}%
}
\end{table*}

This work is situated within research on model adaptation and robustness in Split Computing, with a particular focus on practical deployment scenarios where the head and tail networks are provided as closed, non-modifiable components.
Existing studies on Split Computing have primarily improved inference performance through architectural optimization, feature compression, or parameter tuning, targeting specific challenges such as latency reduction, communication constraints, or robustness to packet loss.
In parallel, parameter-efficient adaptation methods such as adapters and LoRA have demonstrated effective adaptation to data or task shifts with limited trainable parameters; however, these approaches inherently rely on access to internal model structures and parameters, which restricts their applicability in closed Split Computing environments.

In contrast, SALT is designed to adapt intermediate representations under closed-model constraints by explicitly targeting three types of distortions that arise in practical Split Computing deployments.
Specifically, SALT addresses (i) user-specific data distribution shifts, (ii) communication-induced degradation of intermediate representations, and (iii) representation distortions caused by external mechanisms applied for practical privacy protection, such as noise injection.
Rather than modifying the head or tail networks, SALT performs adaptation by applying a lightweight correction directly to intermediate representations, while preserving their dimensionality and compatibility with existing Split Computing architectures and communication protocols.
Importantly, SALT does not aim to provide privacy protection by itself.
Instead, privacy-preserving mechanisms are treated as external sources of representation distortion, and SALT focuses on mitigating the resulting performance degradation through adaptation.

By unifying adaptation to data heterogeneity, communication degradation, and privacy-induced distortion within a single lightweight framework, SALT fills a gap between architectural optimization, parameter-efficient adaptation, and robustness-oriented approaches, enabling reliable and high-accuracy Split Computing in realistic closed-model deployment scenarios.

Table~\ref{tab:rw_compact} summarizes the positioning of SALT relative to existing studies in Split Computing, highlighting differences in their primary objectives, adaptation targets, and assumptions regarding model accessibility.

\section{SALT: SPLIT-ADAPTIVE LIGHTWEIGHT TUNING}

\label{sec:method}

\subsection{System Model}
\label{subsec:system_model}

\subsubsection{Split Computing Architecture}
We consider a standard split computing architecture for edge--cloud collaborative inference.
A deep neural network is partitioned into an edge device-side head network and a server-side tail network. 
The head network, denoted by $H(\cdot)$, is deployed on a resource-constrained client edge device, while the tail network, denoted by $T(\cdot)$, is executed on a cloud or edge server.

Given an input sample $\mathbf{x}$, the client computes an intermediate representation (latent feature)
\begin{equation}
    \mathbf{z} = H(\mathbf{x}),
\end{equation}
which is transmitted to the server over a communication channel.
The server then produces the final prediction
\begin{equation}
    \hat{\mathbf{y}} = T(\mathbf{z}),
\end{equation}
where $\hat{\mathbf{y}}$ denotes the output logits or class probabilities.

The split point is assumed to be located at an intermediate layer of the original model.
Earlier split points reduce client-side computation but yield low-level features that are closer to the raw input, whereas deeper split points increase abstraction at the cost of higher computational load and larger intermediate representations.
Throughout this work, the dimensionality of $\mathbf{z}$ is denoted by $d_z$.

\subsubsection{Closed-Model Constraint}
% Inaccessible head/tail parameters, black-box assumption
We focus on closed-model split computing environments, in which the internal parameters and architectures of both the head and tail networks are inaccessible to the user.
Specifically, we assume that the parameters of $H(\cdot)$ and $T(\cdot)$ are fixed and cannot be modified, fine-tuned, or retrained.

This assumption reflects practical deployment scenarios in real-world edge--cloud systems, such as proprietary AI services accessed through restricted APIs, pre-trained models embedded in hardware accelerators, or security-sensitive systems where model internals are intentionally hidden to protect intellectual property.
In such environments, conventional personalization approaches that rely on updating model parameters (e.g., retraining or fine-tuning the head network) are infeasible.

Formally, during both inference and adaptation phases, the mappings $H(\cdot)$ and $T(\cdot)$ are treated as black-box functions.
The only accessible information is the input--output behavior: the client can observe $\mathbf{z} = H(\mathbf{x})$, and the server can compute the loss $\mathcal{L}(\hat{\mathbf{y}}, \mathbf{y})$ based on the output $\hat{\mathbf{y}} = T(\mathbf{z})$, where $\mathbf{y}$ denotes the ground-truth label.
No gradients with respect to the internal parameters of $H(\cdot)$ or $T(\cdot)$ can be applied to update these networks.

As a consequence, model adaptation must be achieved without altering the head or tail networks and without increasing the dimensionality of the transmitted latent feature.
This constraint fundamentally differentiates the considered problem from conventional split learning and motivates the design of lightweight, external adaptation mechanisms.

\subsection{Problem Formulation}
\label{problem_formulation}
In this subsection, we formulate the problem of user-specific adaptation in split computing under closed-model constraints.

We consider a pre-trained split model consisting of an edge device-side head network $H(\cdot)$ and a server-side tail network $T(\cdot)$, as defined in Section~\ref{subsec:system_model}.
The parameters and internal structures of both $H(\cdot)$ and $T(\cdot)$ are fixed and inaccessible.
Each user is associated with a user-specific data distribution $\mathcal{D}_u$ that may differ from the distribution used to pre-train the original model.

Given a training sample $(\mathbf{x}, \mathbf{y}) \sim \mathcal{D}_u$, the prediction is obtained as
\begin{equation}
    \hat{\mathbf{y}} = T(\mathbf{z}), \quad \mathbf{z} = H(\mathbf{x}),
\end{equation}
and the inference performance is measured by a task loss function
$\mathcal{L}(\hat{\mathbf{y}}, \mathbf{y})$, such as cross-entropy loss for classification tasks.

The objective of user-specific adaptation is to minimize the expected task loss over
$\mathcal{D}_u$:
\begin{equation}
    \min \; \mathbb{E}_{(\mathbf{x}, \mathbf{y}) \sim \mathcal{D}_u}
    \left[ \mathcal{L}\big(T(H(\mathbf{x})), \mathbf{y}\big) \right].
\end{equation}
However, unlike conventional fine-tuning or retraining approaches, direct optimization of the parameters of $H(\cdot)$ and $T(\cdot)$ is not permitted due to the closed-model constraint.

In addition to data distribution shifts, we consider practical edge--cloud environments where the intermediate representation $\mathbf{z}$ may be degraded or perturbed before being processed by the server.
Such degradation may arise not only from unintentional communication impairments, such as packet loss or random feature dropping, but also from intentional perturbations applied for privacy protection.
A simple and widely adopted example of the latter is noise injection into the intermediate representation to mitigate information leakage.

Let $\tilde{\mathbf{z}}$ denote the feature received at the server after such
degradation or perturbation, modeled as
\begin{equation}
    \tilde{\mathbf{z}} = \mathcal{C}(\mathbf{z}),
\end{equation}
where $\mathcal{C}(\cdot)$ represents a generic degradation or perturbation process,
including communication impairments and intentional noise injection.
The prediction under degraded or perturbed representations is then given by
\begin{equation}
    \hat{\mathbf{y}} = T(\tilde{\mathbf{z}}).
\end{equation}

Under these conditions, the problem addressed in this work is to design a lightweight adaptation mechanism for closed split computing that enables user-specific inference while being robust to communication degradation.

Formally, the adaptation mechanism should minimize the expected task loss
\begin{equation}
    \min \; \mathbb{E}_{(\mathbf{x}, \mathbf{y})\sim \mathcal{D}_u}
    \left[ \mathcal{L}\big(T(\mathcal{C}(\phi(\mathbf{z}))), \mathbf{y}\big) \right],
\end{equation}
where $\phi(\cdot)$ denotes an adaptation mechanism applied to the intermediate representation.
This optimization is subject to the constraints that: (i) the parameters of the head and tail networks $H(\cdot)$ and $T(\cdot)$ remain fixed, and (ii) the adaptation mechanism does not introduce additional communication overhead.

\subsubsection{Overview of SALT}
\label{subsec:overview_of_salt}

\begin{figure}[t]
\centering
    \includegraphics[width=\linewidth]{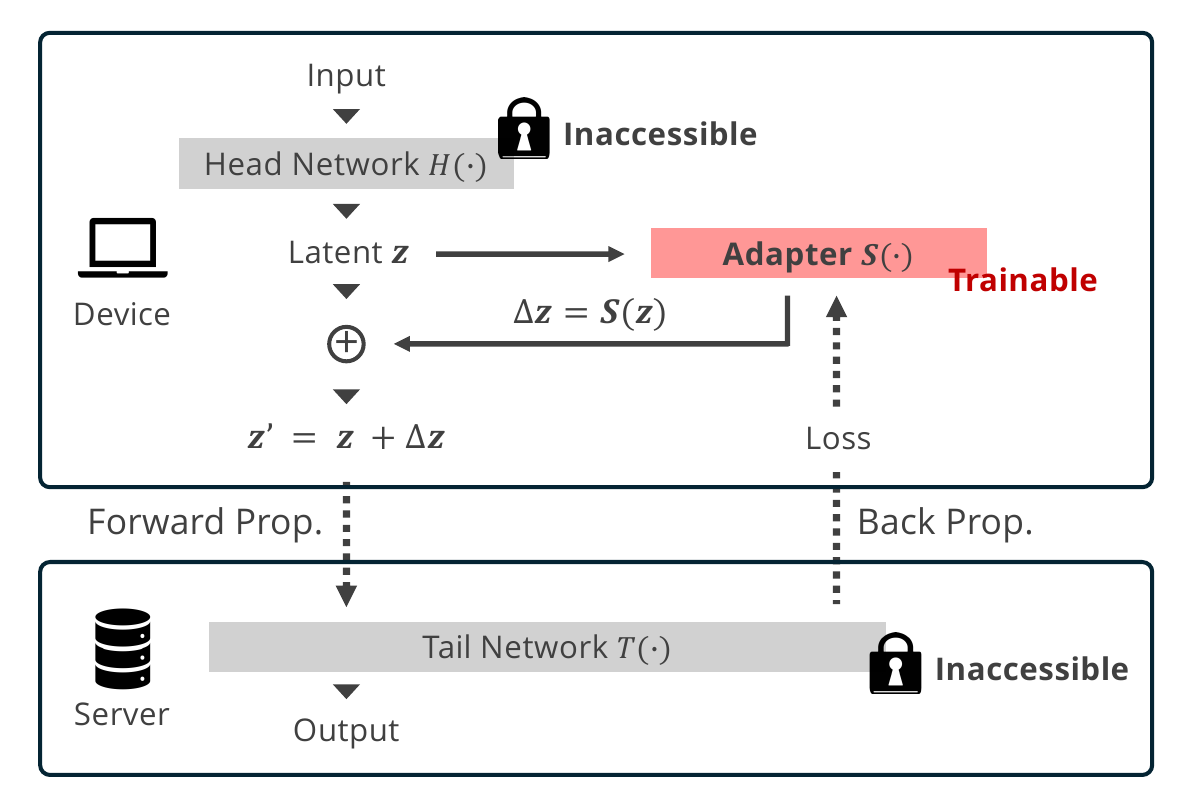}
    \caption{Architecture of SALT, where a trainable adapter applies a correction to the latent feature produced by a frozen head network. The adapted feature is perturbed by noise and forwarded to a frozen tail network on the server. Only the adapter parameters are updated during training.}
    \label{fig:architecture}
\end{figure}

% Motivation and high-level idea of SALT
To address the problem formulated in Section~\ref{problem_formulation}, we propose SALT (Split-Adaptive Lightweight Tuning), a lightweight adaptation framework designed for closed split computing environments.

The key idea of SALT is to realize user-specific adaptation by operating directly on the intermediate representation produced by the frozen head network, without modifying the parameters of either the head or tail network.
Specifically, SALT instantiates the abstract adaptation mechanism $\phi(\cdot)$ introduced in the problem formulation as a compact, trainable module inserted between the head network $H(\cdot)$ and the tail network $T(\cdot)$ on the client-side.

An overview of the SALT architecture and its training setting is illustrated in Fig.~\ref{fig:architecture}.
Given an input sample $\mathbf{x}$, the client computes the intermediate representation $\mathbf{z} = H(\mathbf{x})$ and transmits a refined feature obtained by the SALT module to the server, where the frozen tail network performs the final inference.

SALT provides a unified framework for adaptation under multiple practical objectives, without requiring any modification to the underlying method itself.
In all cases, SALT adapts the intermediate representation while keeping the head and tail networks fixed; the difference lies solely in the adaptation objective used during training.

In this work, we consider three representative adaptation objectives.
The first is adaptation to user-specific data distributions, where SALT is trained to improve inference accuracy for a particular user domain.
The second is adaptation to communication degradation, where SALT is trained under simulated packet loss or unreliable network conditions to enhance robustness to feature corruption during transmission.
The third is adaptation to noise perturbations applied to the intermediate representation, which can be used as a simple and practical mechanism for privacy protection.
Importantly, these objectives are not mutually exclusive.
SALT can be trained with any single objective or with a combination of multiple objectives by appropriately defining the training conditions.
As a result, SALT serves as a flexible adaptation framework that supports user personalization, communication robustness, and privacy-aware inference in a unified manner.

\subsection{Adapter Architecture}
\label{subsec:adapter_architecture}
% Structure of the lightweight adapter module
In this section, we describe the architecture of the adapter module used in SALT.
The adapter serves as a lightweight, trainable component that realizes the abstract adaptation mechanism $\phi(\cdot)$ introduced in the problem formulation, while satisfying the constraints of closed split computing.

% ↓ どちらがいいかな...
% \subsection{Design principles}
% The adapter architecture is designed according to the following principles.
% First, the adapter must operate independently of the internal parameters and structures of the head and tail networks, as these networks are assumed to be fixed and inaccessible.
% Second, the adapter must preserve the dimensionality of the intermediate representation so that no additional communication overhead is introduced.
% Third, the adapter should be computationally lightweight to enable efficient training and inference on resource-constrained client devices.

\subsubsection{Design principles}
The adapter architecture is designed according to the following principles:
\begin{itemize}
    \item The adapter must operate independently of the internal parameters
    and structures of the head and tail networks, as these networks are
    assumed to be fixed and inaccessible.
    \item The adapter must preserve the dimensionality of the intermediate
    representation so that no additional communication overhead is introduced.
    \item The adapter should be computationally lightweight to enable efficient
    training and inference on resource-constrained client devices.
\end{itemize}

\subsubsection{Adapter placement}
The adapter is inserted between the frozen head network $H(\cdot)$ and the frozen tail network $T(\cdot)$.
Given an input sample $\mathbf{x}$, the head network produces an intermediate representation $\mathbf{z} = H(\mathbf{x})$, which is then processed by the adapter before transmission to the server.
The adapted representation is subsequently fed into the tail network for final inference.

In addition, placing the adapter on the client-side allows feature adaptation to be performed before transmission.
This design choice is motivated by two key considerations.
First, client-side placement enables adaptation to be performed on a per-client basis, allowing the intermediate representation to be adapted to client-specific conditions, including both local data distributions and communication environments.
Second, the adapter is assumed to reside locally on the client and is not accessible to external adversaries.
As a result, potential attackers observing transmitted intermediate representations cannot access or incorporate the client-side adapter into their inversion process, limiting reconstruction attacks to noise-perturbed features without adaptation.

\subsubsection{Residual adapter formulation}
In this work, we adopt a residual-style adapter as the primary architectural instantiation of the SALT framework.
The residual formulation adapts the intermediate representation by estimating a correction relative to the original feature, rather than replacing it entirely.

Formally, given the intermediate representation $\mathbf{z} = H(\mathbf{x})$, the adapted representation $\mathbf{z}'$ is defined as
\begin{equation}
    \mathbf{z}' = \mathbf{z} + S(\mathbf{z}),
\end{equation}
where $S(\cdot)$ denotes a lightweight, trainable adapter network.
By construction, this formulation preserves the dimensionality of the intermediate representation, i.e., $\dim(\mathbf{z}') = \dim(\mathbf{z})$, and therefore does not introduce additional communication overhead.

This residual design is particularly suitable for closed split computing environments, as it preserves the original intermediate representation and enables identity-preserving behavior when only minimal adaptation is required.
When the optimal adaptation is close to the identity mapping, the output of the adapter naturally converges toward zero, and the overall behavior reduces to that of the original split computing model.

\subsubsection{Network structure}
The adapter network $S(\cdot)$ is implemented as a compact convolutional neural network operating on the intermediate representation.
In our experiments, $S(\cdot)$ is instantiated as a small number of convolutional layers followed by batch normalization and non-linear activation functions.
The number of channels is kept identical to that of the input representation, and no spatial downsampling or upsampling is performed.
As a result, the adapter preserves both the spatial resolution and the channel dimensionality of the intermediate representation.

While a specific instantiation is used in the evaluation, the SALT framework itself does not depend on a particular adapter structure.
More generally, the architecture and capacity of the adapter can be regarded as design parameters, which may be determined in a parameterized or automated manner, in line with prior studies on architecture optimization.

\subsubsection{Generality of the architecture}
While a residual-style adapter is adopted as the primary instantiation in this work, the SALT framework itself is not restricted to this specific design.
More generally, SALT defines a class of lightweight adaptation mechanisms that operate on the intermediate representation between the head and tail networks under closed-model constraints.

Any architecture that processes the intermediate representation and preserves its dimensionality can be used to instantiate the adaptation mechanism $\phi(\cdot)$.
For example, insertion-style adapters that append a compact trainable module after the head network can also satisfy the design principles of SALT.
These alternative instantiations differ in architectural form but share the same underlying objective of adapting the intermediate representation without modifying the head or tail networks.
A quantitative comparison between the two adapter types is provided in the performance evaluation.

% ここまでOK

\subsection{Training Procedure}
\label{subsec:training_procedure}

\begin{algorithm}[t]
\caption{Training procedure based on split learning}
\label{alg:salt_training}
\begin{algorithmic}[1]
\REQUIRE Frozen head network $H(\cdot)$, Frozen tail network $T(\cdot)$, Trainable SALT module $S(\cdot)$, Training dataset $\mathcal{D} = \{(x_k, y_k)\}$
\FOR{each epoch}
    \FOR{each mini-batch $B = \{(x_k, y_k)\}$ in $\mathcal{D}$}
        \STATE \textbf{Client-side:}
        \STATE \hspace{1.5em} $Z \leftarrow \{ H(x_k) \mid x_k \in B \}$
        \STATE \hspace{1.5em} $\Delta Z \leftarrow \{ S(z_k) \mid z_k \in Z \}$
        \STATE \hspace{1.5em} $Z' \leftarrow \{ z_k + \Delta z_k \mid z_k \in Z, \Delta z_k \in \Delta Z \}$
        \STATE \hspace{1.5em} Transmit $\tilde{Z}'$ to server
        \STATE \textbf{Server-side:}
        \STATE \hspace{1.5em} Receive $\tilde{Z}'= C_{\mathrm{c \rightarrow s}}(Z')$
        \STATE \hspace{1.5em} $\hat{Y} \leftarrow \{ T(\tilde{z}'_k) \mid \tilde{z}'_k \in \tilde{Z}' \}$
        \STATE \hspace{1.5em} $\mathcal{L} \leftarrow \frac{1}{|B|} \sum_{x_k \in B} \text{Cross-Entropy}(\hat{y}_k, y_k)$
        \STATE \hspace{1.5em} $\delta \leftarrow$ gradients of $\mathcal{L}$ w.r.t. $\tilde{Z}'$
        \STATE \hspace{1.5em} Transmit $\delta$ to client
        \STATE \textbf{Client-side:}
        \STATE \hspace{1.5em} Receive $\tilde{\delta}= C_{\mathrm{s \rightarrow c}}(\delta)$
        \STATE \hspace{1.5em} Update parameters of $S(\cdot)$ using $\tilde{\delta}$
    \ENDFOR
\ENDFOR
\end{algorithmic}
\end{algorithm}
This section describes the training procedure of SALT under closed split computing constraints.
The training follows a split learning paradigm, where the head network and the SALT adapter are executed on the client-side, while the tail network is executed on the server-side, as summarized in Algorithm~\ref{alg:salt_training}.

\subsubsection{Training setting}
During training, the parameters of the head network $H(\cdot)$ and the tail network $T(\cdot)$ are fixed and remain unchanged.
Only the parameters of the SALT adapter $S(\cdot)$ are optimized.
This setting reflects closed-model environments, where the internal parameters of deployed models are inaccessible and cannot be modified.

Let $\mathcal{D} = \{(\mathbf{x}_k, \mathbf{y}_k)\}$ denote the training dataset.
For each mini-batch $B \subset \mathcal{D}$, the client computes the intermediate representation $\mathbf{z}_k = H(\mathbf{x}_k)$ for each input sample.
The adapter then processes the intermediate representation to obtain an adapted feature, which is transmitted to the server for further processing.

\subsubsection{Forward and backward propagation}
As shown in Algorithm~\ref{alg:salt_training}, the forward and backward passes are performed collaboratively between the client and the server.
On the client-side, the intermediate representations produced by the frozen head network are adapted by the trainable SALT module.
The adapted features are transmitted to the server through a communication channel, which may introduce degradation such as packet loss or noise.

On the server-side, the received features are processed by the frozen tail network to produce the prediction.
The task loss is computed using a standard loss function $\mathcal{L}(\hat{\mathbf{y}}, \mathbf{y})$, such as cross-entropy loss for classification tasks.
Gradients of the loss with respect to the received intermediate features are then computed and transmitted back to the client.
Upon receiving the gradients, the client updates only the parameters of the adapter network using backpropagation.
Throughout the training process, no gradients are applied to the head or tail networks.

\subsubsection{Training conditions and adaptation objectives}
A key property of SALT is that different adaptation objectives are realized through the training conditions, rather than by modifying the model architecture or the loss function.
In all cases, the same task loss is used, and the underlying training procedure remains unchanged.

Specifically, different operating conditions are incorporated by defining the channel operators
$C_{\mathrm{c \rightarrow s}}(\cdot)$ and
$C_{\mathrm{s \rightarrow c}}(\cdot)$ in
Algorithm~\ref{alg:salt_training}.
These operators model transformations applied to the intermediate representation and the backpropagated gradients during training.

In this work, we consider the following representative training conditions:
\begin{itemize}
    \item \textbf{User-specific adaptation}: The adapter is trained using
    samples drawn from a user-specific data distribution, enabling
    adaptation to data distribution shifts.
    \item \textbf{Communication robustness}: The adapter is trained with
    simulated communication impairments, such as random feature
    dropping, to improve robustness against packet loss and unreliable
    networks.
    \item \textbf{Noise-robust adaptation}: The adapter is trained with
    noise injected into the intermediate representation, allowing it to
    compensate for accuracy degradation caused by noise-based privacy
    protection.
\end{itemize}

These training conditions can be applied individually or in combination.
As a result, SALT provides a unified training framework that supports adaptation to heterogeneous user data distributions, communication degradation, and noise perturbations without altering the model architecture or the optimization objective.

\subsection{Inference procedure}
After training, the adapter network is deployed on the client edge device together with the frozen head network.
During inference, the client computes the intermediate representation, applies the trained adapter, and transmits the adapted feature to the server.
The server performs inference using the frozen tail network without any modification.
Thus, the inference pipeline of SALT is fully compatible with existing split computing systems and introduces no additional computation or architectural changes on the server-side.

\section{PERFORMANCE EVALUATION}

\label{sec:evaluation}

\subsection{Experimental Setup}
\label{subsec:experimental_setup}
This subsection describes the experimental setup used to evaluate the effectiveness of SALT under various operating conditions.
%We follow the general experimental protocol established in our prior work presented at \textit{IEEE GLOBECOM Workshop 2025}, and extend it to include additional evaluations related to privacy-aware inference.
All experiments are designed to assess adaptation performance under closed split computing constraints.

\subsubsection{Datasets and Tasks}
We evaluate SALT using image classification tasks on the CIFAR-10 and CIFAR-100 datasets~\cite{cifar10}.
CIFAR-10 consists of $60{,}000$ color images across $10$ classes, while CIFAR-100 contains the same number of images but is organized into $100$ fine-grained classes, resulting in a more challenging classification task.
For both datasets, $50{,}000$ images are used for training and $10{,}000$ images for testing.
All images are standardized to have zero mean and unit variance before being fed into the network.

Image classification accuracy is used as the primary performance metric throughout the evaluation.
This task setting is chosen because intermediate representations in convolutional neural networks for image classification are known to retain substantial visual information, making them suitable for evaluating both robustness and privacy-related properties in split computing.

While the backbone model is trained on the full datasets, user-specific adaptation scenarios considered in this work are defined by restricting the task to a subset of classes, as described in the following subsection.

\subsubsection{Backbone Model}

\begin{figure}[t]
    \centering
    \includegraphics[width=\linewidth]{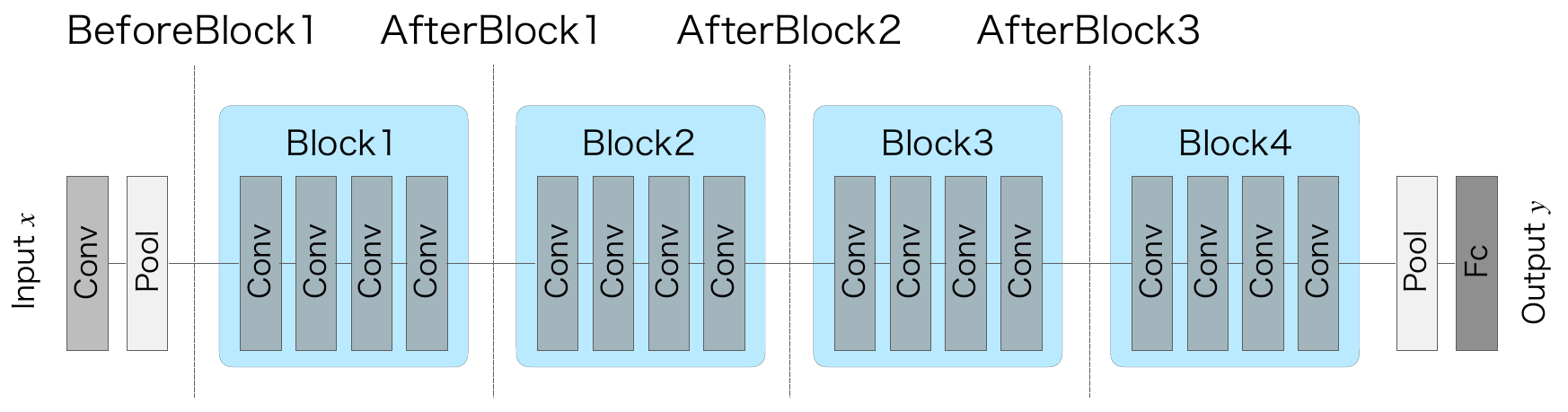}
    \caption{Definition of split points in ResNet-18.
    The head network consists of the layers before the selected split
    point, while the tail network includes the remaining layers.
    We consider four split locations: BeforeBlock1, AfterBlock1,
    AfterBlock2, and AfterBlock3.}
    \label{fig:split_points}
\end{figure}

The backbone model used in our experiments is a ResNet-based convolutional neural network, following the architecture commonly adopted in split computing studies~\cite{resnet}.
The network is divided into a head network deployed on the client-side and a tail network deployed on the server-side.
Multiple split points are considered to evaluate the impact of the split location on adaptation performance, as illustrated in Fig.~\ref{fig:split_points}.

The backbone models are pre-trained on the full training datasets and serve as fixed feature extractors in all experiments.
For CIFAR-10, the model is trained on all 10 classes and achieves a top-1 accuracy of $89.4\%$ on the standard 10-class classification task.
For CIFAR-100, the model is trained on all 100 classes and achieves a top-1 accuracy of $65.7\%$ on the corresponding 100-class classification task.
These pre-trained models are used as the backbone for all subsequent experiments.

In all experiments, the head and tail networks are kept fixed, and adaptation is performed solely through the SALT adapter inserted between them.

\subsubsection{Adapter Configuration}
In the experimental evaluation, the SALT adapter is instantiated as a lightweight three-layer convolutional neural network.
Specifically, the adapter first applies a $3\times3$ convolutional layer, followed by batch normalization and a ReLU activation function, without changing the number of channels.
This sequence is repeated twice, and a final $1\times1$ convolutional layer is applied to produce the correction output $\Delta \mathbf{z}$.
Throughout the adapter, the number of channels is kept identical to that of the input intermediate representation; for example, $128$ channels are maintained at the split point AfterBlock2.

This configuration is used consistently across all experiments unless otherwise stated.
We emphasize that this represents one concrete instantiation of the SALT framework for evaluation purposes, and the framework itself is not restricted to this specific adapter structure.

In terms of model complexity, the adapter introduces a modest overhead compared to the backbone head network.
In our implementation, the adapter contains approximately $443\,\mathrm{K}$ parameters and requires $28.39\,\mathrm{MMACs}$ per forward pass.
By comparison, the head network, defined as the ResNet-18 backbone up to the selected split point, contains between $1.73\,\mathrm{K}$ and $2.78\,\mathrm{M}$ parameters depending on the split configuration.
For example, at the deepest split point considered in our experiments (AfterBlock3), the head network consists of layers from conv1 through layer3, and contains approximately $2.78\,\mathrm{M}$ parameters and $54.72\,\mathrm{MMACs}$.

\subsubsection{Hyperparameters}

\begin{table}[t]
    \centering
    \caption{Training configuration and environment used for all experiments.}
    \label{tab:training_config}
    \resizebox{\linewidth}{!}{%
    \begin{tabular}{|c |c |}
    \hline
    \textbf{Configuration Item} & \textbf{Details} \\
    \hline
    Optimizer           & Adam \\
    Loss Function       & Cross Entropy Loss (on tail network outputs) \\
    Learning Rate       & $1\mathrm{e}{-3}$ \\
    Batch Size          & $128$ \\
    Training Epochs     & $100$ (or until validation loss plateaus) \\
    Activation Function & ReLU \\
    Pooling Type        & Max pooling \\
    Standardization     & Input standardized (mean 0, std 1) \\
    Split Points        & BeforeBlock1, AfterBlock1, AfterBlock2, AfterBlock3 \\
    Packet Loss Rates   & $0.0, 0.25, 0.5, 0.75$ \\
    Noise Distribution  & Zero-mean Gaussian \\
    Noise Strength ($\sigma$) & $0.0$--$1.5$ \\
    Network Bandwidth   & $80~\mathrm{Mbit/s}$ \\
    GPU Environment     & NVIDIA Quadro RTX 6000 \\
    \hline
    \end{tabular}
    }
    \vspace{-2mm}
\end{table}

All experiments are conducted using a common set of hyperparameters to ensure fair comparison across different adaptation objectives and baselines.
The training configuration and computational environment are summarized in Table~\ref{tab:training_config}.

The Adam optimizer is used to train the SALT adapter with a fixed learning rate~\cite{adam}.
Cross-entropy loss computed at the output of the tail network is used as the task loss for all experiments.
Unless otherwise specified, models are trained for 100 epochs or until the validation loss plateaus.

In addition to communication degradation modeled by packet loss, we also consider intentional perturbations to intermediate representations.
Specifically, zero-mean Gaussian noise with varying standard deviation $\sigma$ is applied to the intermediate features during training and inference when evaluating robustness against privacy-related perturbations.

\subsection{Operating Conditions and Assumptions}
\label{subsubsec:operating_conditions}

\subsubsection{User-Specific Data Distribution}

User-specific data distributions are modeled following the experimental protocol used in our prior study, where a backbone model pre-trained on a complete dataset is adapted to a user-defined domain by restricting the available data to a subset of classes.
This setting reflects realistic edge–cloud scenarios in which users observe only a limited portion of the original task space.

\begin{figure}[t]
\centering
    \includegraphics[width=\linewidth]{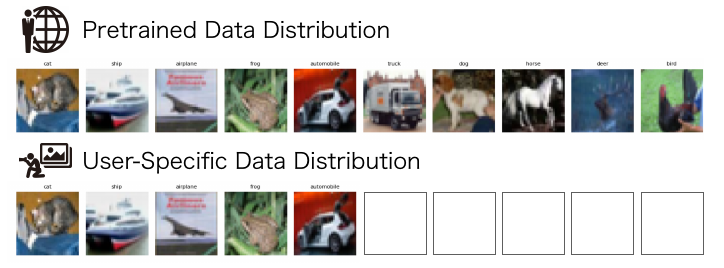}
    \caption{Illustration of the adaptation setting. While the model is pre-trained on a complete dataset, only a subset of these classes is used to simulate user-specific data.}
    \label{fig:adaptation_setting}
\end{figure}

For CIFAR-10, the backbone model is pre-trained on all ten classes using the standard training set.
As illustrated in Fig.~\ref{fig:adaptation_setting}, a user-specific task is constructed by selecting only five classes (\textit{airplane, automobile, bird, cat, deer}), corresponding to labels 0--4, from the original dataset.
This results in a reduced dataset consisting of $25{,}000$ training images and $5{,}000$ test images.
All adaptation methods, including SALT and baseline approaches, are trained and evaluated exclusively on this subset.

For CIFAR-100, we follow the same class-restriction principle while considering a more challenging and data-scarce setting.
The backbone model is pre-trained on the full dataset with $100$ classes, and user-specific adaptation is evaluated on a subset of five classes.
Since CIFAR-100 contains only $500$ images per class, this results in a significantly smaller amount of user-specific training data compared to CIFAR-10.
This setting enables us to examine whether SALT can effectively adapt to user-specific data distributions under a fine-grained classification scenario with limited labeled data.

During adaptation, the parameters of the head and tail networks remain fixed, and only the SALT adapter is trained using samples drawn from the user-specific dataset.
Evaluation is performed on the corresponding user-specific test set, allowing us to isolate the effect of adaptation to user-specific data distributions under closed-model constraints.

\subsubsection{Communication Degradation Model}
% Packet loss, feature dropping, lossy channels
Communication between the client and server is modeled following the packet-loss-aware split computing framework used in prior work.
We consider both feature degradation due to packet loss and communication latency caused by limited network bandwidth.

\paragraph{Packet loss model}
Let $\mathbf{z} \in \mathbb{R}^{C \times H \times W}$ denote the intermediate representation produced by the head network.
Packet loss is modeled as element-wise random feature dropping using a probabilistic binary mask.
Specifically, the received feature $\tilde{\mathbf{z}}$ is given by
\begin{equation}
    \tilde{\mathbf{z}} = \mathbf{z} \odot \mathbf{m}(p),
\end{equation}
where $\odot$ denotes the Hadamard product and $\mathbf{m}(p) \in \{0,1\}^{C \times H \times W}$ is a binary mask whose elements are independently sampled from a Bernoulli distribution with parameter $1 - p$.
Here, $p \in \{0.0, 0.25, 0.5, 0.75\}$ denotes the packet loss rate.

Following prior work, missing feature elements caused by packet loss are zero-filled during forward propagation.
The same packet loss model is applied during both training and inference when evaluating robustness to communication degradation.

\paragraph{Communication latency model}
Communication latency is estimated numerically based on the size of the transmitted intermediate representation and the available network bandwidth.
Assuming that intermediate features are transmitted during both the forward and backward passes in split learning, the per-batch communication latency is computed as
\begin{equation}
    T_{\mathrm{comm/batch}} = 2 \times \frac{V_{\mathbf{z}} \cdot B}{\theta},
\end{equation}
where $V_{\mathbf{z}}$ denotes the size of the intermediate representation in bits, $B$ is the batch size, and $\theta$ represents the network bandwidth.
In all experiments, the bandwidth is fixed at $80~\mathrm{Mbit/s}$.

The total communication latency during training is then given by
\begin{equation}
    T_{\mathrm{comm,total}} =
    E \times \left\lceil \frac{N}{B} \right\rceil \times
    T_{\mathrm{comm/batch}},
\end{equation}
where $E$ is the number of training epochs determined by early stopping, and $N$ is the number of training samples.
This formulation allows us to quantify the impact of communication overhead on training efficiency under realistic edge--cloud deployment conditions.

\subsubsection{Adversarial Observation Assumption}
We consider a privacy threat model in which an adversary attempts to reconstruct input images from intermediate representations transmitted from the client to the server.
Specifically, we focus on an inversion attack scenario, where the attacker observes the intermediate features exchanged during split computing and aims to recover the original input data.

The attacker is assumed to have knowledge of the backbone model architecture and to train an inversion model using a surrogate dataset drawn from the same distribution as the training data, following a commonly adopted white-box assumption in the inversion attack literature.
However, the attacker does not have access to the original input samples used during inference, nor to the client-side SALT adapter, which is locally deployed and remains inaccessible.

We emphasize that the goal of this threat model is not to provide provable privacy guarantees against all possible adversaries, but to evaluate the practical effectiveness of noise injection as a simple and widely used privacy protection mechanism in split computing.
Under this setting, we assess whether SALT can maintain high inference accuracy while preserving the privacy benefits introduced by noise-based defenses.

\subsection{Baselines}

We also consider a variant of the proposed method, referred to as SALT (Insertion), which represents an alternative design choice within the SALT framework.
In SALT (Insertion), trainable layers are inserted after the split point to adapt intermediate representations, while keeping the original head and tail networks fixed.
This variant satisfies the closed-model constraint and enables user-specific adaptation in a manner consistent with SALT.

\begin{table}[t]
\centering
\caption{Comparison of adaptation strategies in Split Computing.}
\label{tab:baseline_comparison}
\resizebox{\linewidth}{!}{%
\begin{tabular}{|c|cccc|}
\hline
Method & Adaptation Target & Parameter Access & Training Cost & Closed-Model Applicable \\
\hline
No Adaptation & None & -- & None & \checkmark \\
Head Retraining & Head Network & Full & High & $\times$ \\
Head Fine-Tuning & Head Network & Partial & Medium & $\times$ \\
SALT (Insertion) & Latent Features & Local Only & Medium--Low & \checkmark \\
SALT (Adapter)   & Latent Features & Local Only & Low & \checkmark \\
\hline
\end{tabular}%
}
\end{table}

As summarized in Table~\ref{tab:baseline_comparison}, both SALT (Insertion) and SALT (Adapter) satisfy the closed-model constraint and enable user-specific adaptation without modifying the original model parameters.
While both approaches share the same design philosophy, they differ in how the adaptation module interacts with the intermediate representations.
Due to these structural differences, SALT (Insertion) may require more training epochs to converge, potentially resulting in higher adaptation cost compared to SALT (Adapter).

\subsection{Results}

\subsubsection{Adaptation to User-Specific Data Distributions}
\label{subsec:eval_user_data}
We first evaluate the effectiveness of SALT in adapting to user-specific data distributions under closed split computing constraints, where the head and tail networks remain fixed and only a lightweight adapter module is trained.

\begin{table*}[t]
  \centering
  \small
  \caption{Performance Comparison of Adaptation Methods on CIFAR-10 and CIFAR-100}
  \label{tab:total_results}
  \resizebox{\linewidth}{!}{%
  \begin{tabular}{|c |cccc|cccc|}
    \hline
    \multirow{2}{*}{Method} & \multicolumn{4}{c|}{CIFAR-10} & \multicolumn{4}{c|}{CIFAR-100} \\
    \cline{2-9}
    & Accuracy & Epochs & Comp./Comm. Latency [s] & Inf. Latency [s] & Accuracy & Epochs & Comp./Comm. Latency [s] & Inf. Latency [s] \\
    \hline
    Original & $0.881$ & - & - / - & $5.3$ & $0.669$ & - & - / - & $1.0$ \\
    Retrain & $0.924$ & $54.0$ & $857.7$ / $1110.2$ & $5.4$ & $0.824$ & $49.6$ & $122.1$ / $208.3$ & $1.0$ \\
    Tune & $0.920$ & $26.0$ & $372.3$ / $534.6$ & $5.4$ & $0.889$ & $39.6$ & $101.3$ / $166.3$ & $1.0$ \\
    SALT (Insertion Adapter) & $0.934$ & $25.3$ & $287.6$ / $520.2$ & $5.3$ & $0.902$ & $27.2$ & $60.7$ / $114.2$ & $1.0$ \\
    SALT (Residual Adapter) & $0.938$ & $20.5$ & $260.6$ / $421.5$ & $5.3$ & $0.902$ & $21.2$ & $54.0$ / $89.0$ & $1.0$ \\
    \hline
  \end{tabular}
  }
\end{table*}

\begin{figure}[t]
    \centerline{\includegraphics[width=\linewidth]{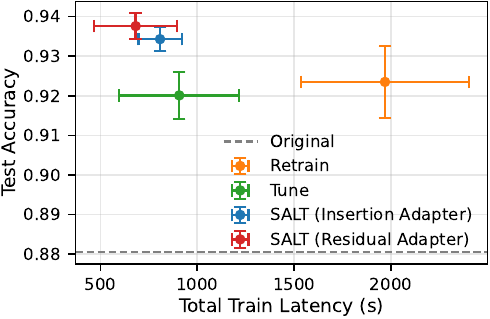}}
    \caption{Accuracy vs. Total Training Latency for each method on CIFAR-10. SALT maintains high accuracy with low training cost. Error bars indicate $95\%$ confidence intervals over 10 trials.}
    \label{fig:cifar10_scatter}
\end{figure}

\begin{figure}[t]
    \centerline{\includegraphics[width=\linewidth]{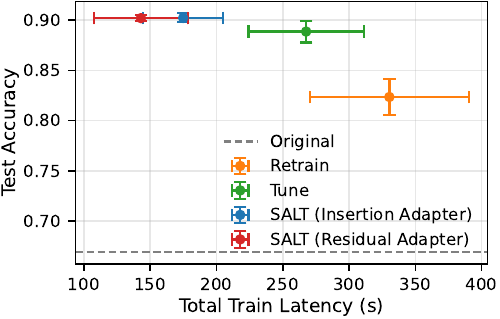}}
    \caption{Accuracy vs. Total Training Latency for each method on CIFAR-100. SALT maintains high accuracy with low training cost. }
    \label{fig:cifar100_scatter}
\end{figure}

Figures~\ref{fig:cifar10_scatter} and~\ref{fig:cifar100_scatter} show the relationship between test accuracy and total training latency for different adaptation methods on CIFAR-10 and CIFAR-100, respectively.
In these plots, the horizontal axis represents the total training latency, while the vertical axis denotes the classification accuracy.
Points located in the upper-left region indicate better performance, corresponding to higher accuracy achieved with lower training cost.
The baseline labeled Original corresponds to the pre-trained split model without any adaptation.

On CIFAR-10 (Fig.~\ref{fig:cifar10_scatter}), retraining achieves relatively high accuracy ($92.4\%$) but incurs the largest training cost, requiring $54.0$ epochs and over $1{,}900$ seconds of total computation and communication latency.
Fine-tuning reduces the training cost to $26.0$ epochs, but results in a lower accuracy of $92.0\%$.
In contrast, SALT attains the highest classification accuracy among the compared methods while requiring substantially less training time.
Specifically, SALT (Residual Adapter) achieves an accuracy of $93.8\%$ with only $20.5$ training epochs, reducing the total training latency by approximately $65\%$ compared to retraining.
These results demonstrate that SALT achieves a favorable accuracy--efficiency trade-off by adapting the intermediate representations without modifying the head network parameters.

We further compare two design variants within the SALT framework, namely SALT (Adapter) and SALT (Insertion).
While both variants improve the accuracy--latency trade-off compared to conventional baselines, SALT (Adapter) consistently achieves comparable or higher accuracy with lower training cost.
For example, on CIFAR-10, SALT (Residual Adapter) converges in $20.5$ epochs, whereas SALT (Insertion Adapter) requires $25.3$ epochs to reach a similar accuracy level.
This difference can be attributed to their adaptation mechanisms.
SALT (Adapter) is trained to output a corrective vector that adjusts the transmitted latent features, where a zero-valued correction is sufficient when the pre-trained model already matches the target condition.
In contrast, SALT (Insertion) learns to generate transformed intermediate representations themselves, which generally requires learning a larger functional mapping and may therefore lead to slower convergence.

A similar trend is observed on CIFAR-100 (Fig.~\ref{fig:cifar100_scatter}), where SALT variants tend to occupy the upper-left region of the plot, achieving higher accuracy with lower training latency compared to conventional baselines.
In this more challenging classification task, the performance gap between SALT and retraining becomes more evident, suggesting that the efficiency advantage of SALT is maintained as the complexity of the data distribution increases.

\begin{figure}[t]
    \centerline{\includegraphics[width=\linewidth]{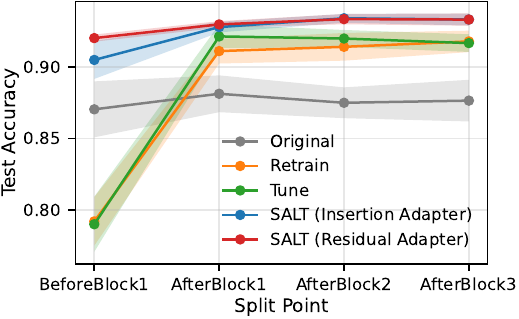}}
    \caption{Accuracy comparison across different split points on CIFAR-10. SALT consistently outperforms baseline methods regardless of split depth, achieving stable personalization without accessing or fine-tuning the head network.}
    \label{fig:splitpoint_accuracy}
\end{figure}

Figure~\ref{fig:splitpoint_accuracy} further examines the impact of the split location on adaptation performance using CIFAR-10.
Across all considered split points, SALT maintains stable and high classification accuracy.
This observation indicates that the effectiveness of SALT is not sensitive to the choice of split point, which is particularly important in practical split computing systems where the split location may be constrained by edge device capabilities or system design considerations.
This robustness to split-point selection can be attributed to the design of SALT, which operates by adapting intermediate representations through lightweight, localized corrections rather than relying on modifications to the upstream feature extraction process.
Since SALT directly compensates for distributional and communication-induced mismatches at the representation level, its effectiveness is less dependent on the depth or semantic abstraction of the features at the split point.

Overall, these results confirm that SALT enables effective user-specific adaptation under closed-model constraints, achieving up to a $65\%$ reduction in training cost while improving classification accuracy compared to conventional retraining and fine-tuning approaches.

\subsubsection{Robustness to Communication Degradation}
\label{subsec:eval_comm_degradation}
% Packet loss experiments

\begin{figure}[t]
    \centerline{\includegraphics[width=\linewidth]{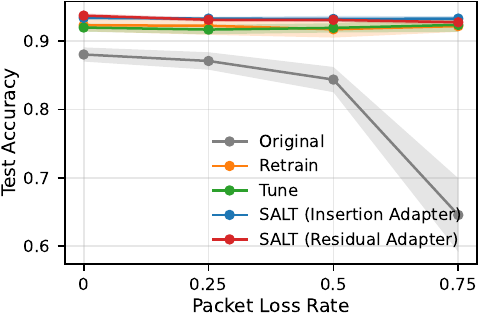}}
    \caption{Accuracy under varying packet loss rates on CIFAR-10. While the original split model collapses under severe packet loss, SALT maintains stable and high accuracy across all loss rates without accessing the head or tail network parameters.}
    \label{fig:lossrate_accuracy}
\end{figure}

\begin{figure}[t]
    \centerline{\includegraphics[width=\linewidth]{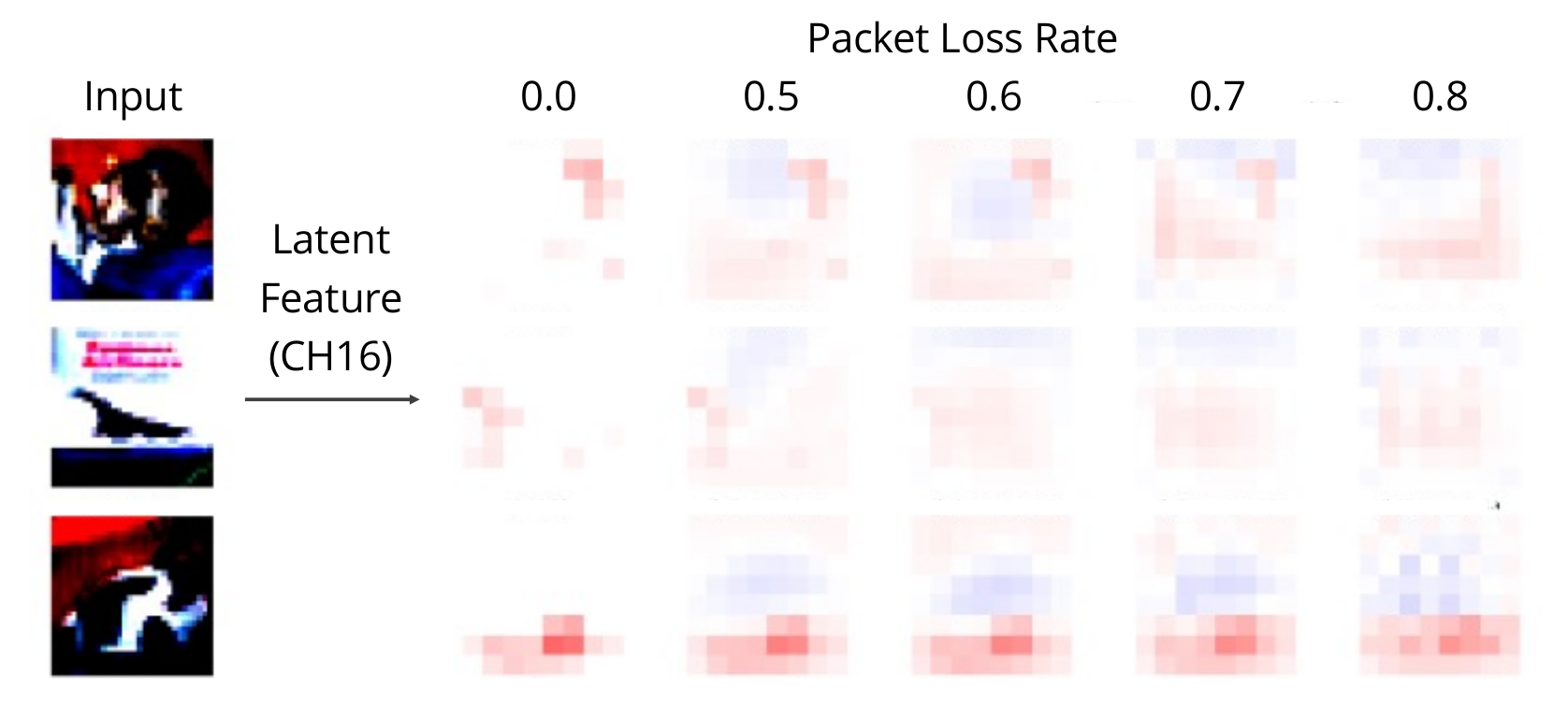}}
    \caption{Intermediate representations before packet loss under different loss rates. The figure shows channel-wise latent features (CH16) before feature dropping, for models trained with different packet loss rates. Higher loss rates lead to more spatially distributed representations, indicating redundancy-aware feature learning that improves robustness to packet loss.}
    \label{fig:pr_illustration}
\end{figure}

We next evaluate the robustness of SALT against communication degradation in split computing, focusing on packet loss during the transmission of intermediate representations.

Figure~\ref{fig:lossrate_accuracy} reports the inference accuracy on CIFAR-10 under varying packet loss rates.
The horizontal axis represents the packet loss rate, which ranges from $0.0$ (lossless transmission) to $0.75$, while the vertical axis denotes the classification accuracy of the main task.
The gray curve corresponds to the original split model without adaptation.

As the packet loss rate increases, the accuracy of the original model degrades monotonically and collapses beyond approximately $50\%$ packet loss, indicating strong sensitivity to missing feature elements.
Fine-tuning and retraining partially improve robustness, but their performance still deteriorates significantly under severe packet loss.
In contrast, SALT consistently maintains higher classification accuracy across all packet loss rates.
Even under high-loss conditions with a packet loss rate of $75\%$, SALT preserves over $90\%$ accuracy, demonstrating strong robustness to severe network degradation.
These results indicate that SALT can adapt the intermediate representation not only to user-specific data distributions but also to the communication environment assumed during training.

To gain insight into why SALT is robust to packet loss, Fig.~\ref{fig:pr_illustration} visualizes representative intermediate representations under different loss conditions.
The figure shows channel-wise outputs (CH16) of the intermediate layer for the same input samples, where the packet loss rate is varied from $0.0$ to the range of $0.5$--$0.8$.
For clarity, the visualized feature maps correspond to the intermediate representations before packet dropping, while packet loss is applied during training to simulate information loss and to enable robustness learning.

In non-robust models (e.g., the original split model and models trained under low-loss conditions), the intermediate representation tends to rely on localized and sparse features.
In such cases, critical information is concentrated in specific spatial regions, and losing these regions due to packet dropping leads to severe degradation in inference performance.
By contrast, SALT trained under packet loss conditions learns more distributed feature representations, where task-relevant information is spread across a broader spatial region rather than being encoded in a small number of critical feature elements.
As a result, even when a subset of feature elements is lost during transmission, the remaining features retain sufficient semantic information to support accurate inference.
This qualitative observation explains the robustness trends observed in Fig.~\ref{fig:lossrate_accuracy} and highlights the effectiveness of SALT for robust split computing over unreliable communication channels.

\subsubsection{Robustness to Noise Injection for Privacy-Aware Inference}
\label{subsec:eval_noise}

\begin{figure}[t]
    \centerline{\includegraphics[width=\linewidth]{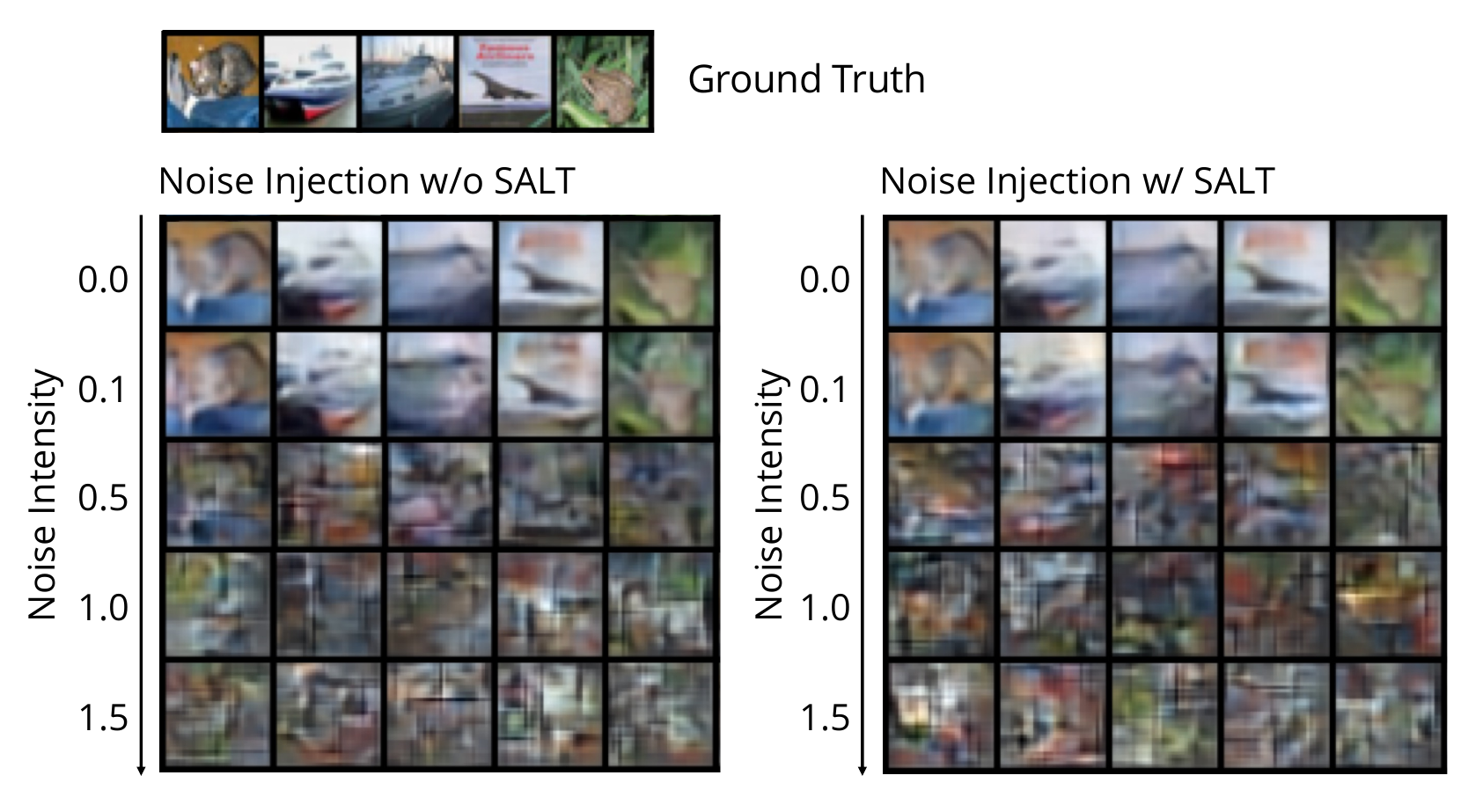}}
    \caption{Visual comparison of inversion attack results under noise injection with and without SALT. The attacker is trained on noise-free representations and evaluated on noisy ones; reconstruction quality degrades similarly in both cases, indicating that SALT does not weaken noise-based privacy protection.}
    \label{fig:ni_illustration}
\end{figure}

\begin{figure}[t]
    \centerline{\includegraphics[width=0.85\linewidth]{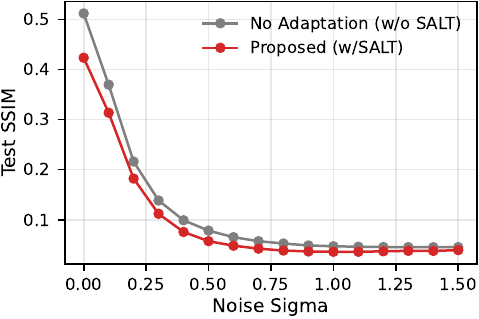}}
    \caption{Effect of noise strength on inversion attack performance (SSIM). SSIM decreases rapidly with increasing noise strength, and the nearly identical trends confirm that SALT does not degrade noise-based privacy defenses.}
    \label{fig:sigma_ssim}
\end{figure}

\begin{figure}[t]
    \centerline{\includegraphics[width=0.85\linewidth]{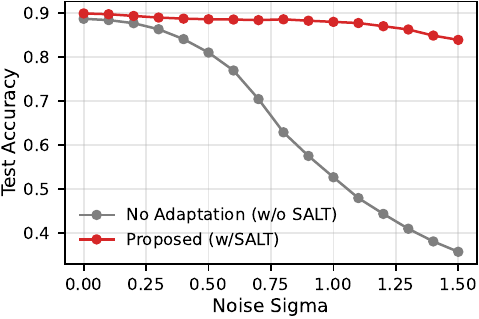}}
    \caption{Effect of noise strength on main-task classification accuracy. While noise injection severely degrades accuracy without adaptation, SALT maintains high accuracy across a wide range of noise strengths, demonstrating robustness to privacy-oriented perturbations.}
    \label{fig:sigma_acc}
\end{figure}

We finally evaluate the effectiveness of SALT under a privacy-aware inference setting in which noise injection is employed as a simple and practical mechanism to mitigate information leakage from intermediate representations.
Unless otherwise stated, the experiments in this subsection are conducted using the residual-type adapter, which serves as the primary architectural instantiation of SALT.
In this experiment, random noise sampled from a zero-mean Gaussian distribution is added to the intermediate features during both training and inference, and the noise strength is controlled by the standard deviation $\sigma$.
All evaluations are conducted on the full CIFAR-10 dataset.

It is important to emphasize that SALT itself does not provide privacy protection.
Instead, privacy is achieved through noise injection, which inevitably degrades inference accuracy.
The role of SALT in this setting is to compensate for the accuracy loss caused by noise injection, thereby enabling high task performance while preserving the privacy benefits introduced by the noise-based defense.

Figure~\ref{fig:ni_illustration} provides a qualitative overview of the impact of noise injection on inversion attacks.
The figure visualizes reconstructed images obtained by an inversion attacker trained using noise-free intermediate representations.
As the noise strength increases, reconstructed images become progressively distorted and eventually visually unrecognizable in both cases, demonstrating that stronger noise effectively enhances privacy by hindering input reconstruction.
Importantly, the comparable degradation trends indicate that introducing SALT does not diminish the effectiveness of noise injection as a privacy protection mechanism.

Figure~\ref{fig:sigma_ssim} quantitatively evaluates this effect using the structural similarity index (SSIM)~\cite{ssim}, which measures the structural resemblance between the reconstructed and original images, with higher values indicating greater visual similarity.
As $\sigma$ increases, the SSIM value decreases rapidly for both methods, dropping from approximately $0.5$ at $\sigma=0.0$ to below $0.1$ at $\sigma=0.5$, and converging to around $0.04$ for $\sigma \geq 1.0$.
The nearly overlapping SSIM curves with and without SALT indicate that SALT does not weaken the privacy protection effect introduced by noise injection.

Figure~\ref{fig:sigma_acc} shows the corresponding main-task classification accuracy under the same noise conditions.
Without SALT, inference accuracy degrades sharply as the noise strength increases, falling from approximately $89\%$ at $\sigma=0.0$ to below $50\%$ at $\sigma=1.1$, and further to around $35\%$ at $\sigma=1.5$.
In contrast, SALT maintains high accuracy across a wide range of noise strengths, preserving approximately $88\%$ accuracy at $\sigma=1.0$ and remaining above $84\%$ even at $\sigma=1.5$.
These results demonstrate that SALT effectively compensates for the accuracy degradation caused by noise injection while preserving the privacy benefits achieved by the noise-based defense.

It is worth noting that SALT is deployed on the client-side and is not accessible to the attacker.
The attacker is assumed to have access only to the noisy intermediate representations and the architecture of the head network, but not to the structure or parameters of the SALT module.
Under this realistic threat model, the above results demonstrate that SALT enables a favorable accuracy--privacy trade-off, achieving strong resistance to inversion attacks while maintaining high inference performance under noise injection.

\section{CONCLUSION}
\label{sec:conclusion}
This work proposed SALT, a lightweight adaptation framework for split computing under closed-model constraints, where the head and tail networks are fixed and inaccessible. By introducing a compact client-side adapter that refines intermediate representations, SALT enables effective model adaptation without modifying the original models or increasing communication overhead. Extensive experiments on CIFAR-10 and CIFAR-100 with ResNet-18 demonstrated that SALT achieves a favorable accuracy–efficiency trade-off across multiple practical operating conditions. In user-specific adaptation on CIFAR-10, SALT improved accuracy from $88.1\%$ to $93.8\%$ while reducing training latency by more than $60\%$ compared to retraining. SALT also maintained high accuracy under severe packet loss (over $90\%$ at a packet loss rate of $75\%$) and preserved robust inference performance under noise injection used for privacy protection.

These results indicate that SALT provides a practical and unified adaptation mechanism for real-world split computing systems, enabling robust and efficient inference under data heterogeneity, unreliable communication, and privacy-oriented perturbations without requiring access to model internals. Future work includes extending SALT to larger and more complex models, as well as evaluating its effectiveness in real-world edge–cloud deployment environments. Another important direction is to further investigate the theoretical properties of SALT from a communication-theoretic perspective, including its interaction with feature transmission, channel impairments, and resource constraints in distributed inference systems.

\bibliographystyle{IEEEtran}
\bibliography{references/main}

@inproceedings{neurosurgeon,
  author = {Kang, Yiping and Hauswald, Johann and Gao, Cao and Rovinski, Austin and Mudge, Trevor and Mars, Jason and Tang, Lingjia},
  title = {Neurosurgeon: Collaborative Intelligence Between the Cloud and Mobile Edge},
  year = {2017},
  isbn = {9781450344654},
  publisher = {Association for Computing Machinery},
  address = {New York, NY, USA},
  doi = {10.1145/3037697.3037698},
  booktitle = {Proc. Int. Conf. Archit. Support for Programming Languages and Oper. Syst.},
  pages = {615–629},
  numpages = {15},
  location = {Xi'an, China},
  series = {ASPLOS '17}
}

@article{jointdnn,
  author = {Eshratifar, Amir Erfan and Abrishami, Mohammad Saeed and Pedram, Massoud},
  title = {JointDNN: An Efficient Training and Inference Engine for Intelligent Mobile Cloud Computing Services},
  year = {2021},
  issue_date = {Feb. 2021},
  publisher = {IEEE Educational Activities Department},
  address = {USA},
  volume = {20},
  number = {2},
  issn = {1536-1233},
  doi = {10.1109/TMC.2019.2947893},
  journal = {IEEE Trans. Mobile Comput.},
  month = feb,
  pages = {565–576},
  numpages = {12}
}

@article{sc_ee_survey,
  author = {Matsubara, Yoshitomo and Levorato, Marco and Restuccia, Francesco},
  title = {Split Computing and Early Exiting for Deep Learning Applications: Survey and Research Challenges},
  year = {2022},
  issue_date = {May 2023},
  publisher = {Association for Computing Machinery},
  address = {New York, NY, USA},
  volume = {55},
  number = {5},
  issn = {0360-0300},
  doi = {10.1145/3527155},
  journal = {ACM Comput. Surv.},
  month = dec,
  articleno = {90},
  numpages = {30}
}

@ARTICLE{ib_task_oriented_comm,
  author={Shao, Jiawei and Mao, Yuyi and Zhang, Jun},
  journal={IEEE J. Sel. Areas Commun.}, 
  title={Learning Task-Oriented Communication for Edge Inference: An Information Bottleneck Approach}, 
  year={2022},
  volume={40},
  number={1},
  pages={197-211},
  doi={10.1109/JSAC.2021.3126087}
}

@article{sec_semcom_survey,
  author = {Zhang, Milin and Abdi, Mohammad and Dasari, Venkat R. and Restuccia, Francesco},
  title = {Semantic Edge Computing and Semantic Communications in 6G networks: A unifying survey and research challenges},
  year = {2025},
  issue_date = {Oct 2025},
  publisher = {Elsevier North-Holland, Inc.},
  address = {USA},
  volume = {270},
  number = {C},
  issn = {1389-1286},
  doi = {10.1016/j.comnet.2025.111531},
  journal = {Comput. Netw.},
  month = oct,
  numpages = {21}
}

@inproceedings{fredrikson_mi_confidence,
  author = {Fredrikson, Matt and Jha, Somesh and Ristenpart, Thomas},
  title = {Model Inversion Attacks that Exploit Confidence Information and Basic Countermeasures},
  year = {2015},
  isbn = {9781450338325},
  publisher = {Association for Computing Machinery},
  address = {New York, NY, USA},
  doi = {10.1145/2810103.2813677},
  booktitle = {Proc. ACM SIGSAC Conf. Comput. Commun. Secur. (CCS)},
  pages = {1322–1333},
  numpages = {12},
  location = {Denver, Colorado, USA},
  series = {CCS '15}
}

@inproceedings{refil_fisher_split_privacy,
  title={Measuring and Controlling Split Layer Privacy Leakage Using Fisher Information},
  author={Kiwan Maeng and Chuan Guo and Sanjay Kariyappa and Edward Suh},
  booktitle={Workshop on Federated Learning, NeurIPS},
  year={2022}
}

@inproceedings{delphi_crypto_inference,
  author = {Mishra, Pratyush and Lehmkuhl, Ryan and Srinivasan, Akshayaram and Zheng, Wenting and Popa, Raluca Ada},
  title = {Delphi: A Cryptographic Inference System for Neural Networks},
  year = {2020},
  isbn = {9781450380881},
  publisher = {Association for Computing Machinery},
  address = {New York, NY, USA},
  doi = {10.1145/3411501.3419418},
  booktitle = {Proc. Workshop Privacy-Preserving Mach. Learn. Pract. (PPMLP)},
  pages = {27–30},
  numpages = {4},
  location = {Virtual Event, USA},
  series = {PPMLP'20}
}

@misc{label_privacy_peft_split,
  title={Label Privacy in Split Learning for Large Models with Parameter-Efficient Training}, 
  author={Philip Zmushko and Marat Mansurov and Ruslan Svirschevski and Denis Kuznedelev and Max Ryabinin and Aleksandr Beznosikov},
  year={2024},
  eprint={2412.16669},
  archivePrefix={arXiv},
  primaryClass={cs.LG},
}

@inproceedings{aem_blackbox_domain_adapt,
  author = {Xiao, Siying and Ye, Mao and He, Qichen and Li, Shuaifeng and Tang, Song and Zhu, Xiatian},
  title = {Adversarial Experts Model for Black-box Domain Adaptation},
  year = {2024},
  isbn = {9798400706868},
  publisher = {Association for Computing Machinery},
  address = {New York, NY, USA},
  doi = {10.1145/3664647.3681123},
  booktitle = {Proc. ACM Int. Conf. Multimedia (MM)},
  pages = {8982–8991},
  numpages = {10},
  location = {Melbourne VIC, Australia},
  series = {MM '24}
}

@inproceedings{inverting_gradients,
  author    = {Jonas Geiping and Hartmut Bauermeister and Hannah Dr{\"o}ge and Michael Moeller},
  title     = {Inverting Gradients -- How Easy Is It to Break Privacy in Federated Learning?},
  booktitle = {Adv. Neural Inf. Process. Syst. (NeurIPS)},
  year      = {2020}
}

@inproceedings{tinytl,
  author = {Cai, Han and Gan, Chuang and Zhu, Ligeng and Han, Song},
  booktitle = {Adv. Neural Inf. Process. Syst. (NeurIPS)},
  editor = {H. Larochelle and M. Ranzato and R. Hadsell and M.F. Balcan and H. Lin},
  pages = {11285--11297},
  publisher = {Curran Associates, Inc.},
  title = {TinyTL: Reduce Memory, Not Parameters for Efficient On-Device Learning},
  volume = {33},
  year = {2020}
}

@INPROCEEDINGS {nopeek, 
  author = { Vepakomma, Praneeth and Singh, Abhishek and Gupta, Otkrist and Raskar, Ramesh },
  booktitle = {Proc. IEEE Int. Conf. Data Mining Workshops (ICDMW)},
  title = {{ NoPeek: Information leakage reduction to share activations in distributed deep learning }},
  year = {2020},
  volume = {},
  ISSN = {},
  pages = {933-942},
  doi = {10.1109/ICDMW51313.2020.00134},
  publisher = {IEEE Computer Society},
  address = {Los Alamitos, CA, USA},
  month =Nov
}

@inproceedings{dfil_bound_invertibility,
  title={Bounding the Invertibility of Privacy-preserving Instance Encoding using Fisher Information},
  author={Kiwan Maeng and Chuan Guo and Sanjay Kariyappa and G. Edward Suh},
  booktitle={Adv. Neural Inf. Process. Syst. (NeurIPS)},
  year={2023},
}

@inproceedings{salt_arxiv,
  author    = {Yuya Okada and Takayuki Nishio},
  title     = {SALT: A Lightweight Model Adaptation Method for Closed Split Computing Environments},
  booktitle = {Proc. IEEE Global Commun. Conf. Workshops (GLOBECOM Workshops)},
  year      = {2025}
}

@InProceedings{houlsby_adapters,
  title = 	 {Parameter-Efficient Transfer Learning for {NLP}},
  author =       {Houlsby, Neil and Giurgiu, Andrei and Jastrzebski, Stanislaw and Morrone, Bruna and De Laroussilhe, Quentin and Gesmundo, Andrea and Attariyan, Mona and Gelly, Sylvain},
  booktitle = 	 {Proc. Int. Conf. Mach. Learn. (ICML)},
  pages = 	 {2790--2799},
  year = 	 {2019},
  editor = 	 {Chaudhuri, Kamalika and Salakhutdinov, Ruslan},
  volume = 	 {97},
  series = 	 {Proceedings of Machine Learning Research},
  month = 	 {09--15 Jun},
  publisher =    {PMLR}
}

@inproceedings{residual_adapters_visual_domains,
  author = {Rebuffi, Sylvestre-Alvise and Bilen, Hakan and Vedaldi, Andrea},
  title = {Learning multiple visual domains with residual adapters},
  year = {2017},
  isbn = {9781510860964},
  publisher = {Curran Associates Inc.},
  address = {Red Hook, NY, USA},
  booktitle = {Adv. Neural Inf. Process. Syst. (NeurIPS)},
  pages = {506–516},
  numpages = {11},
  location = {Long Beach, California, USA},
  series = {NIPS'17}
}

@inproceedings{lora,
  author    = {Edward J. Hu and Yelong Shen and Phillip Wallis and Zeyuan Allen-Zhu and Yuanzhi Li and Shean Wang and Lu Wang and Weizhu Chen},
  title     = {LoRA: Low-Rank Adaptation of Large Language Models},
  booktitle = {Proc. Int. Conf. Learn. Represent. (ICLR)},
  year      = {2022}
}

@inproceedings{pockengine,
  author = {Zhu, Ligeng and Hu, Lanxiang and Lin, Ji and Chen, Wei-Ming and Wang, Wei-Chen and Gan, Chuang and Han, Song},
  title = {PockEngine: Sparse and Efficient Fine-tuning in a Pocket},
  year = {2023},
  isbn = {9798400703294},
  publisher = {Association for Computing Machinery},
  address = {New York, NY, USA},
  booktitle = {Proc. 56th Annu. IEEE/ACM Int. Symp. Microarchitecture (MICRO)},
  pages = {1381–1394},
  numpages = {14},
  location = {Toronto, ON, Canada},
  series = {MICRO '23}
}

@ARTICLE{comtune,
  author={Itahara, Sohei and Nishio, Takayuki and Koda, Yusuke and Yamamoto, Koji},
  journal={IEEE Access}, 
  title={Communication-Oriented Model Fine-Tuning for Packet-Loss Resilient Distributed Inference Under Highly Lossy IoT Networks}, 
  year={2022},
  volume={10},
  number={},
  pages={14969-14979},
  doi={10.1109/ACCESS.2022.3149336}
}

@INPROCEEDINGS{si_nr_packet_loss_tolerant_split,
  author={Itahara, Sohei and Nishio, Takayuki and Yamamoto, Koji},
  booktitle={Proc. IEEE Global Commun. Conf. (GLOBECOM)}, 
  title={Packet-Loss-Tolerant Split Inference for Delay-Sensitive Deep Learning in Lossy Wireless Networks}, 
  year={2021},
  volume={},
  number={},
  pages={1-6},
  doi={10.1109/GLOBECOM46510.2021.9685179}
}

@ARTICLE{loss_adapter,
  author={Hou, Zhangcheng and Ohtsuki, Tomoaki},
  journal={IEEE Internet Things J.}, 
  title={Loss-Adapter: Addressing Network Packet Loss in Distributed Inference for Lossy IoT Environments}, 
  year={2025},
  volume={12},
  number={12},
  pages={22048-22057},
  doi={10.1109/JIOT.2025.3550162}
}

@INPROCEEDINGS{nas_for_split_computing,
  author={Shimizu, Shoma and Nishio, Takayuki and Saito, Shota and Hirose, Yoichi and Yen-Hsiu, Chen and Shirakawa, Shinichi},
  booktitle={Proc. IEEE Global Commun. Conf. Workshops (GLOBECOM Workshops)}, 
  title={Neural Architecture Search for Improving Latency-Accuracy Trade-off in Split Computing}, 
  year={2022},
  volume={},
  number={},
  pages={1864-1870},
  doi={10.1109/GCWkshps56602.2022.10008544}
}

@ARTICLE{supernet_ft_nas_split,
  author={Ogawa, Kohei and Hara, Yuko},
  journal={IEEE Access}, 
  title={Efficient Neural Architecture Search for Split Computing via Supernet Fine-Tuning}, 
  year={2025},
  volume={13},
  number={},
  pages={166127-166139},
  doi={10.1109/ACCESS.2025.3610959}
}

@techreport{cifar10,
  author      = {Alex Krizhevsky},
  title       = {Learning Multiple Layers of Features from Tiny Images},
  institution = {University of Toronto},
  year        = {2009}
}

@INPROCEEDINGS{resnet,
  author={He, Kaiming and Zhang, Xiangyu and Ren, Shaoqing and Sun, Jian},
  booktitle={Proc. IEEE Conf. Comput. Vis. Pattern Recognit. (CVPR)}, 
  title={Deep Residual Learning for Image Recognition}, 
  year={2016},
  volume={},
  number={},
  pages={770-778},
  doi={10.1109/CVPR.2016.90}
}

@inproceedings{adam,
  author    = {Diederik P. Kingma and Jimmy Ba},
  title     = {Adam: A Method for Stochastic Optimization},
  booktitle = {Proc. Int. Conf. Learn. Represent. (ICLR)},
  year      = {2015}
}

@article{ssim,
  author  = {Zhou Wang and Alan C. Bovik and Hamid R. Sheikh and Eero P. Simoncelli},
  title   = {Image Quality Assessment: From Error Visibility to Structural Similarity},
  journal = {IEEE Trans. Image Process.},
  volume  = {13},
  number  = {4},
  pages   = {600--612},
  year    = {2004}
}

\begin{IEEEbiography}[{\includegraphics[width=1in,height=1.25in,clip,keepaspectratio]{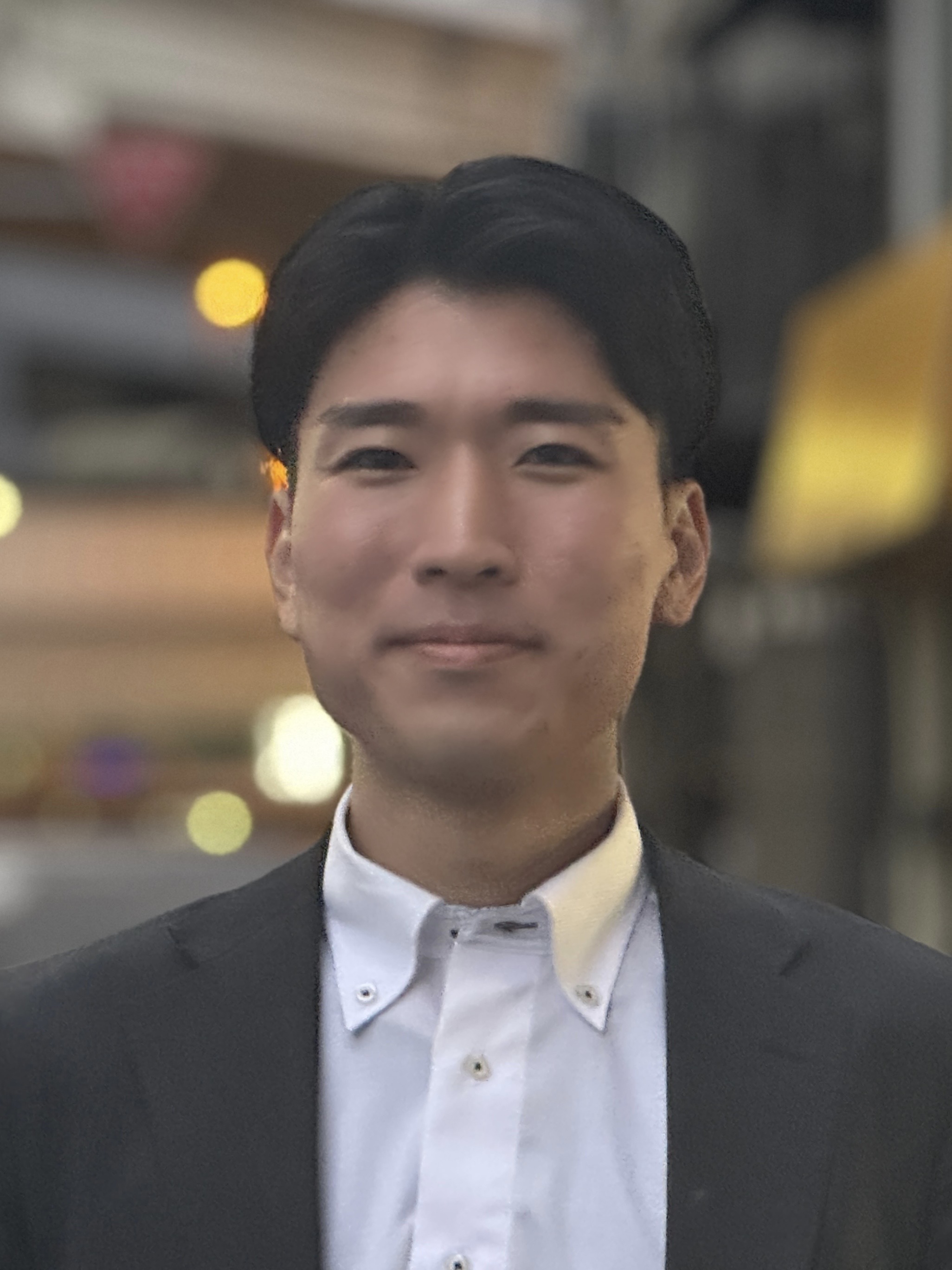}}]{Yuya Okada}~received the B.E. degree in mathematical and systems engineering from Shizuoka University, Japan, in 2024, and conducted research in wireless communications. He is currently pursuing the M.E. degree in information and communications engineering at the School of Engineering, Institute of Science Tokyo, Tokyo, Japan. His current research focuses on split computing for secure and lightweight edge–cloud collaborative inference, particularly in CNN-based image analysis. His research interests include privacy-preserving machine learning, edge AI, and collaborative intelligence.
\end{IEEEbiography}

\begin{IEEEbiography}[{\includegraphics[width=1in,height=1.25in,clip,keepaspectratio]{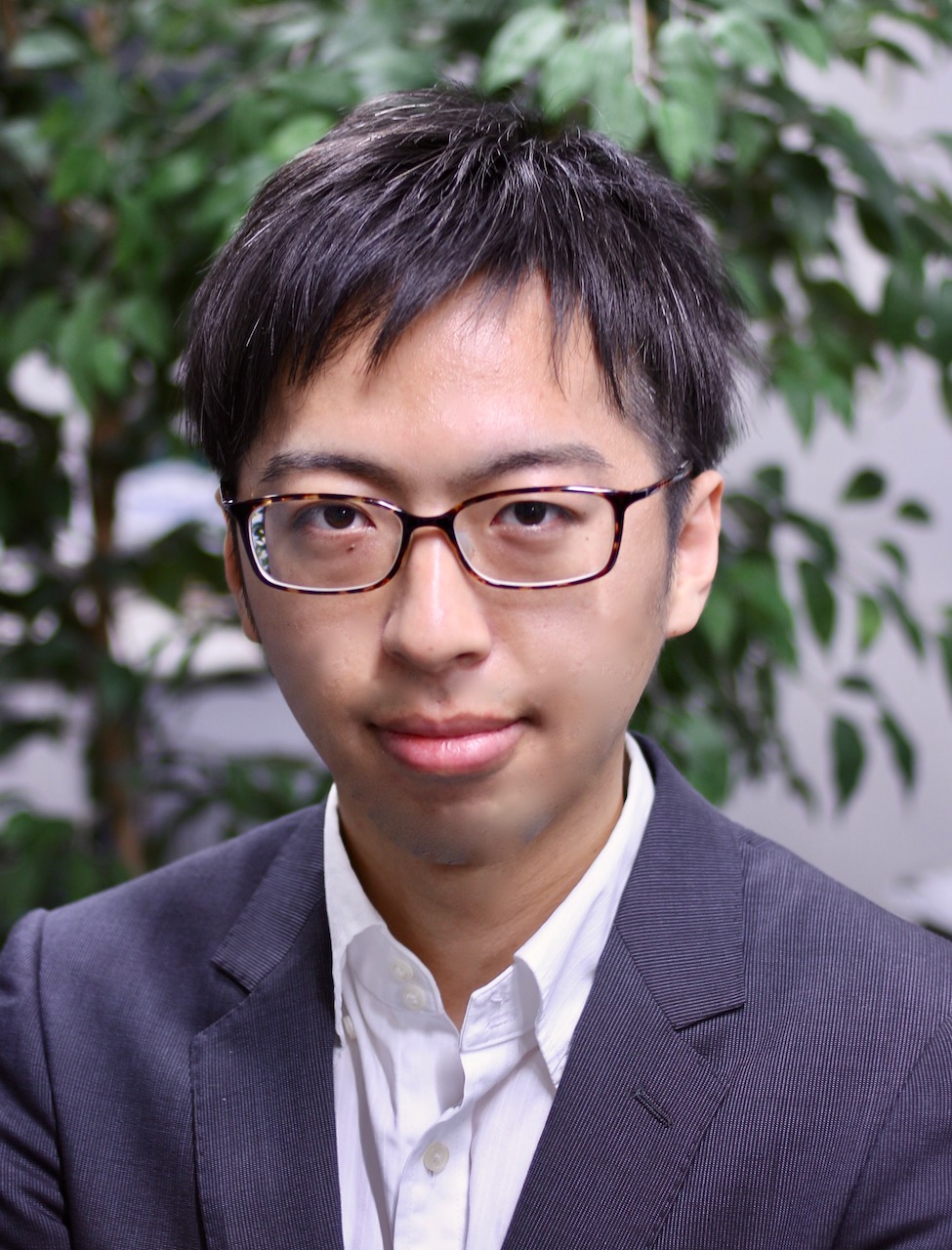}}]{Takayuki Nishio}~(S’11-M’14-SM'20) received the B.E.\ degree in electrical and electronic engineering and the master’s and Ph.D.\ degrees in informatics from Kyoto University in 2010, 2012, and 2013, respectively. He had been an assistant professor in the Graduate School of Informatics, Kyoto University from 2013 to 2020. From 2016 to 2017, he was a visiting researcher in Wireless Information Network Laboratory (WINLAB), Rutgers University, United States. Since 2020, he has been an associate professor at the School of Engineering, Institute of Science Tokyo (formerly Tokyo Institute of Technology), Japan. His current research interests include machine learning-based network control, machine learning in wireless networks, and heterogeneous resource management.
\end{IEEEbiography}

\end{document}